\documentclass[11pt]{article}

\usepackage[preprint]{acl}

\usepackage{times}
\usepackage{latexsym}

\usepackage[T1]{fontenc}

\usepackage[utf8]{inputenc}

\usepackage{microtype}

\usepackage{inconsolata}

\usepackage{graphicx}

\usepackage{hyperref}       
\usepackage{url}            
\usepackage{booktabs}       
\usepackage{amsfonts}       
\usepackage{nicefrac}       
\usepackage{xcolor}         
\usepackage[most]{tcolorbox}
\usepackage{cleveref} 
\usepackage{multirow}
\usepackage{placeins}
\usepackage{colortbl}
\usepackage{siunitx}
\usepackage{xspace}
\usepackage{caption}
\usepackage{bm}
\usepackage{algorithm}
\usepackage{algpseudocode}
\usepackage{rotating}
\usepackage{makecell}
\usepackage{longtable}
\usepackage{fontawesome5}

\usepackage[most]{tcolorbox}
\usepackage{xcolor}

\usepackage{amssymb}     
\usepackage{pifont}

\newcommand{\cmark}{\ding{51}}
\newcommand{\xmark}{\ding{55}}

\usepackage{lineno}

\definecolor{darkblue}{rgb}{0, 0, 0.5}
\hypersetup{colorlinks=true, citecolor=darkblue, linkcolor=darkblue, urlcolor=darkblue}

\newcommand{\name}[1]{\textsc{ReVision}\xspace}

\title{\name{}: Scaling Computer-Use Agents via Temporal Visual Redundancy Reduction}

\author{
\textbf{
Amirhossein Abaskohi$^{1}$\thanks{Corresponding author: \texttt{aabaskoh@cs.ubc.ca}. This work was done during the internship at Microsoft Research~(MSR), Vancouver Lab.},
Yuhang He$^{2}$,
Peter West$^{1}$,}\\
\textbf{
\hspace{0.2em}Giuseppe Carenini$^{1}$,
Pranit Chawla$^{2}$,
Vibhav Vineet$^{2}$}\\[0.5em]
$^{1}$University of British Columbia, $^{2}$Microsoft Research
}

\begin{document}
\maketitle


\begin{abstract}
Computer-use agents (CUAs) rely on visual observations of graphical user interfaces, where each screenshot is encoded into a large number of visual tokens. As interaction trajectories grow, the token cost increases rapidly, limiting the amount of history that can be incorporated under fixed context and compute budgets. This has resulted in no or very limited improvement in the performance when using history unlike other domains. We address this inefficiency by introducing \textbf{\name{}}, which is used to train multimodal language models on trajectories where redundant visual patches are removed using a learned patch selector that compares patch representations across consecutive screenshots while preserving spatial structure required by the model. Across three benchmarks, OSWorld, WebTailBench, and AgentNetBench, when processing trajectories with 5 history screenshots using \textsc{Qwen2.5-VL-7B}, \textbf{\name{} reduces token usage by 46\% on average while improving success rate by 3\% over the no drop baseline}. This establishes a clear efficiency gain, enabling agents to process longer trajectories with fewer tokens. With this improved efficiency, we revisit the role of history in CUAs and find that performance continues to improve as more past observations are incorporated when redundancy is removed\footnote{
Code is available on \href{https://github.com/AmirAbaskohi/ReVision}
{{\textcolor{black}{\faGithub}}~GitHub}.
}.  
\end{abstract}
\section{Introduction}
\label{sec:intro}

Multimodal large language models (MLLMs) have enabled agents that interact with graphical user interfaces (GUIs) by combining visual understanding with language-based reasoning~\citep{wang2025opencuaopenfoundationscomputeruse, wang2025ui}. These computer-use agents (CUAs) operate on screenshots to generate grounded actions such as clicks, typing, and navigation for multi-step tasks. Benchmarks such as VisualWebArena~\citep{koh-etal-2024-visualwebarena}, OSWorld~\citep{OSWorld}, and AgentNetBench~\citep{wang2025opencuaopenfoundationscomputeruse} demonstrate their ability to handle complex workflows across web and desktop environments. Most systems rely primarily on the current screenshot, sometimes with limited history~\citep{sager2025comprehensive}, despite many tasks requiring memory of past states or actions. Scaling such long-horizon reasoning remains challenging due to the need to process extended visual trajectories under constrained context budgets~\citep{chen2026iterresearch}.

\begin{figure*}[t]
    \centering
    \includegraphics[width=\linewidth]{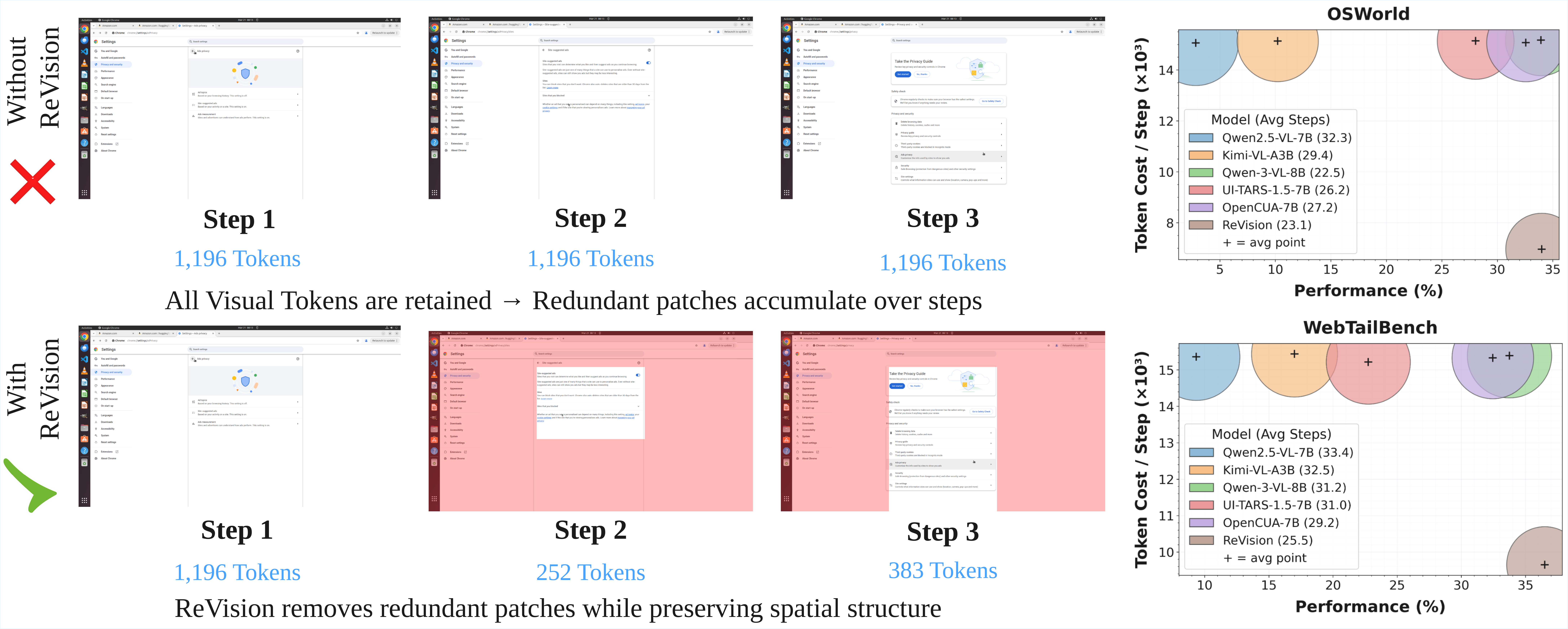}
    \caption{\textbf{Token efficiency with \name{}.} \textbf{Left:} \name{} removes redundant patches across steps, reducing token accumulation while preserving spatial structure. \textbf{Right:} \name{} achieves higher success rates at maximum 100 steps OSWorld and WebTailBench, with lower token cost across models. Circle size indicates average steps to complete tasks.}
    \label{fig:teaser}
\end{figure*}

A straightforward way to provide memory is to append past screenshots to the model context. However, this is highly inefficient: each additional image adds hundreds or thousands of visual tokens, quickly exhausting the context budget. As shown in Figure~\ref{fig:teaser}, much of this cost is redundant because consecutive GUI screenshots largely overlap. The model therefore repeatedly processes unchanged visual content, wasting computation and limiting longer-history reasoning. Reducing this redundancy is not just an efficiency optimization, but a key enabler of better decision-making: it frees context budget for longer, more informative histories that support long-horizon interactions.

To address this inefficiency, we introduce \textbf{\name{}}, a redundancy-aware training framework for MLLMs that operates on trajectories where redundant visual patches are removed across consecutive screenshots. At the core of \name{} is a learned patch selector that compares patch-level representations over time and filters out visually redundant regions while preserving the spatial structure required by the model. Rather than applying token reduction only at inference time, we train the model on filtered trajectories, allowing it to reason over compact visual histories. This enables \name{} to remove redundant visual tokens without architecture changes while remaining compatible with existing MLLMs.

This efficiency improvement directly translates to better performance. Across OSWorld~\citep{OSWorld}, WebTailBench~\citep{awadallah2025fara}, and AgentNetBench~\citep{wang2025opencuaopenfoundationscomputeruse}, when using 5 history screenshots with \textsc{Qwen2.5-VL-7B}~\citep{Qwen2.5-VL}, \textbf{\name{} reduces token usage by approximately 46\% on average while achieving a +3\% gain in success rate over the no-drop baseline}. With only 3 history images, \name{} achieves performance close to the some of the best baseline while using roughly half of the visual tokens. As the history length increases, the gains become more pronounced: with 5 or more images, \name{} consistently outperforms most of the baselines with the same size by at least 2\% on average (Figure~\ref{fig:teaser}, right). \textbf{\name{} shifts the efficiency frontier by enabling longer visual histories under similar compute budgets while achieving higher success rates}.

Our contributions are as follows: \textbf{(i)} we identify and quantify substantial temporal redundancy in sequential screenshots from long computer-use trajectories, showing that a large fraction of visual tokens remains unchanged across consecutive steps; \textbf{(ii)} we introduce \textbf{\name{}}, a \textsc{Qwen2.5-VL-7B} based model trained with a temporal patch scorer that performs patch-level filtering across consecutive screenshots, enabling the model to reason over compact visual histories without modifying the underlying architecture; \textbf{(iii)} we demonstrate across long-horizon computer-use benchmarks that redundancy-aware history filtering reduces token usage, improves success rates, and \textbf{delays the saturation point of visual history}.
\section{Related Work}
\label{sec:relatedwork}

\begin{table*}[h]
    \centering
    \renewcommand{\arraystretch}{1.1}
    \resizebox{\linewidth}{!}{%
    \begin{tabular}{lcccc}
        \toprule
        \textbf{Dataset/Benchmark} 
        & \makecell{\textbf{Avg. Steps}\\\textbf{/ Task}} 
        & \makecell{\textbf{Avg. \# of Patches}\\\textbf{/ Image}} 
        & \makecell{\textbf{Avg. Redundant Patches}\\\textbf{/ Image}} 
        & \makecell{\textbf{Avg. (\%) Redundant Patches}\\\textbf{/ Image}} \\
        \midrule
        AgentNetBench~\citeyear{wang2025opencuaopenfoundationscomputeruse} & 12.1 & 2,284 & 1,014 & 44.4\% \\
        OSWorld~\citeyear{OSWorld} & 16.9 & 2,769 & 1,556 & 56.2\% \\
        WindowsAgentArena~\citeyear{bonatti2024windows} & 11.7 & 2,769 & 1,462 & 52.8\% \\
        WebTailBench~\citeyear{awadallah2025fara} & 22.4 & 2,769 & 1,174 & 42.4\% \\
        Mind2Web2~\citeyear{gou2025mind2web2} & 13.4 & 2,769 & 1,199 & 43.3\% \\
        VisualWebArena~\citeyear{koh2024visualwebarena} & 6.8 & 2,769 & 1,373 & 49.6\% \\
        AndroidWorld~\citeyear{rawles2024androidworlddynamicbenchmarkingenvironment} & 7.6 & 1,196 & 456 & 38.2\% \\
        GUIAct~\citeyear{guicourse2024} & 5.5 & 1,196 & 435.3 & 36.4\% \\
        \midrule
        \textbf{Average} & 12.1 & 2,315 & 1,083 & 45.4\% \\
        \bottomrule
    \end{tabular}
    }
    \caption{\textbf{Dataset-level visual redundancy in computer-use benchmarks.} We report average steps, patches per image, and redundant patches across environments. While \textit{GUIAct} and \textit{AgentNetBench} are offline benchmarks with fixed steps, others depend on agent performance. We use \textsc{GPT-5.4} to ensure minimal and consistent trajectories for fair comparison. Results show that 36\%–56\% of visual tokens are redundant across steps, motivating \textbf{\name{}}.}
    \label{tab:temporal_redundancy}
    \vspace{-1em}
\end{table*}

\noindent
\textbf{Computer-use agents and benchmarks.}
Recent progress in CUAs has been driven by multimodal models that interact with digital environments through screenshots and natural language. Early systems such as WebShop and WebArena rely on structured representations like DOM or accessibility trees~\cite{NEURIPS2022_82ad13ec,zhou2023webarena}. In contrast, a growing line of work adopts a vision-first paradigm, reasoning directly over pixels. Methods such as CogAgent, AGUVIS, OpenCUA, FARA, WebSTAR, and UI-TARS operate purely on visual inputs~\cite{hong2023cogagent,xu2024aguvis, wang2025opencuaopenfoundationscomputeruse, awadallah2025fara, webstar_cua_train, qin2025ui,wang2025ui}. Other approaches, including WebVoyager, SeeAct, and ScaleCUA, incorporate both visual observations and structured signals to improve robustness in complex environments~\cite{he2024webvoyager,zheng2023seeact,liu2025scalecua}. Benchmarks including WebArena, VisualWebArena, OSWorld, and AgentNetBench enable evaluation in long-horizon settings~\cite{zhou2023webarena,koh-etal-2024-visualwebarena,OSWorld,wang2025opencuaopenfoundationscomputeruse}. Despite this progress, agents typically rely on limited visual history, and increasing history length yields diminishing returns, highlighting inefficiencies in naive context scaling~\cite{abhyankar2025osworldhumanbenchmarkingefficiencycomputeruse,kerboua2025focusagent}. Our setting instead requires filtering visual history at patch granularity across consecutive screenshots while preserving temporal evidence for long-horizon decisions.

\noindent
\textbf{Visual token pruning and context compression.}
Prior work reduces visual token usage either within images or across trajectory steps. Methods such as ShowUI and FocusUI prune spatially redundant regions within a single screenshot~\cite{lin2024showui,ouyang2026focusui}, while approaches like Focus-Scan-Refine and adaptive compression further remove tokens based on saliency or importance~\cite{tong2026focus,huang2026ppe,huang2026nwa}. At the trajectory level, methods such as FocusAgent reduce the number of past steps included in context~\cite{kerboua2025focusagent}. However, these approaches operate either spatially within images or temporally at the step level, without explicitly modeling redundancy across consecutive screenshots, leading to repeated processing of unchanged visual regions.

\noindent
\textbf{Temporal redundancy in sequential visual data.}
Temporal redundancy has been extensively studied in video understanding, where consecutive frames share similar content. Prior work addresses this via keyframe selection, feature reuse, and token compression~\cite{choudhury2024don,korbar2019scsampler,choi2024vid,tao2025dycoke,timechatonline}. However, CUAs differ: screenshots evolve through agent actions and must be processed jointly with textual reasoning. Existing approaches operate at the frame or feature level within vision models, whereas our setting requires patch-level filtering in the token space of multimodal LLMs while preserving temporally distributed evidence for long-horizon decision making.
\section{Temporal Visual Redundancy}
\label{sec:redundancy_analysis}

CUAs operate over sequences of screenshots that capture the evolving state of a digital environment. At each step, the model encodes the current screenshot into a large number of visual tokens and processes them together with accumulated textual context to predict the next action. However, consecutive screenshots in a trajectory often exhibit substantial visual overlap: in most steps, only a small region of the interface changes (e.g., a button click or text update), while the majority of the screen remains unchanged. Despite this, standard multimodal models process each image independently, resulting in repeated encoding and consumption of nearly identical visual tokens across time.

To quantify this, we analyze pairs of consecutive screenshots $(I_{t-1}, I_t)$ across multiple benchmarks and measure redundancy by comparing corresponding patches. As shown in Table~\ref{tab:temporal_redundancy}, redundancy is consistently high, with an average of \textbf{45.4\%} of patches unchanged across steps and over \textbf{56\%} in some settings. This corresponds to over \textbf{1,000 redundant patches per step} on average. These findings show that a large portion of computation is spent on repeated visual content and that the context budget is dominated by redundant tokens, limiting the model’s ability to incorporate useful history. This motivates \name{}, which removes redundant visual tokens across time while preserving task-relevant information.

\section{\name{}}
\label{sec:method}

\begin{figure*}[t]
    \centering
    \includegraphics[width=\textwidth]{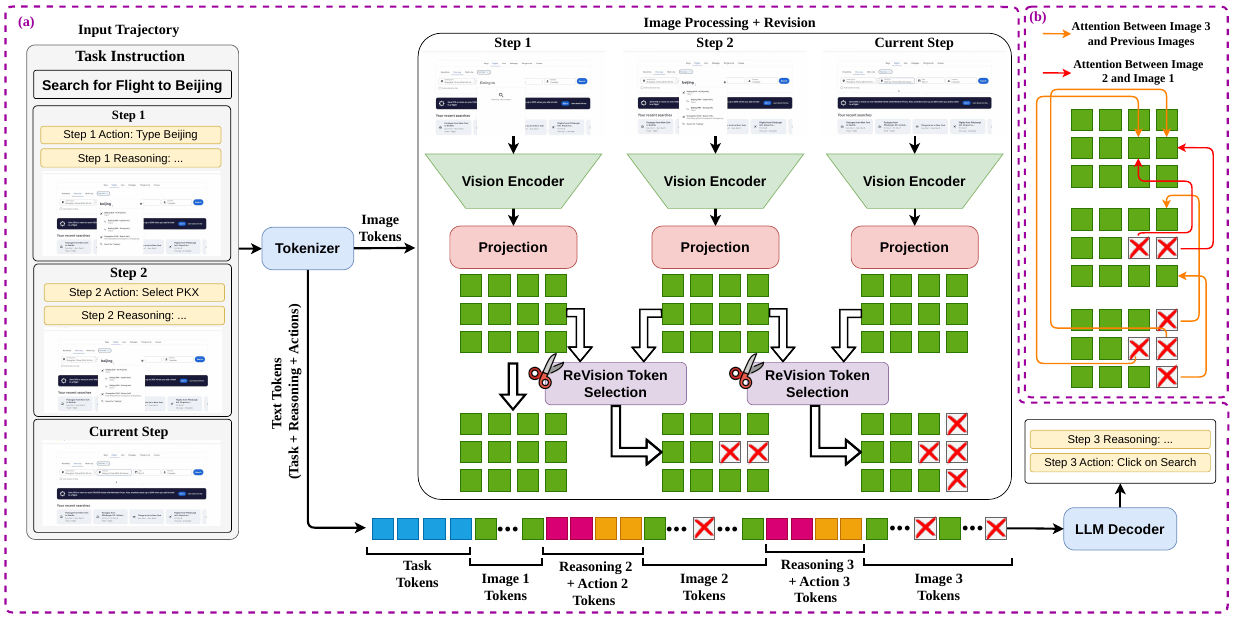}
    \caption{\textbf{Overview of \name{}.} \textbf{(a)} \name{} removes redundant patches by comparing corresponding tokens across consecutive screenshots, reducing visual tokens while preserving spatial alignment before passing them to the LLM. \textbf{(b)} The model learns to attend to relevant regions in previous images, enabling effective reasoning with reduced visual input.}
    \label{fig:method}
    \vspace{-1em}
\end{figure*}

As illustrated in Figure~\ref{fig:method}, \name{} reduces redundant visual tokens across sequential GUI observations by learning to selectively retain only informative patches. Our approach consists of two main components. \textbf{First}, we train a lightweight \textbf{three-layer MLP classifier}, referred to as \textit{ReVision Token Selection} (RTS), which takes as input the embeddings of two corresponding patches from consecutive screenshots and predicts whether the patch in the current image is redundant given the previous one. \textbf{Second}, we integrate RTS into the pipeline of a MLLM and fine-tune the model on AgentNet~\citep{wang2025opencuaopenfoundationscomputeruse} trajectories (with a fixed history image window) where redundant patches are removed from all but the first image. This training setup encourages the model to recover omitted visual information from earlier observations, enabling efficient use of longer histories.

\subsection{Problem Formulation}  
\label{sec:problem_formulation}
CUAs operate over trajectories $\{(I_t, T_t, A_t)\}_{t=1}^T$, where $I_t$ is the screenshot, $T_t$ is the accumulated textual context (all action and reasoning from the preiovus steps and the task description), and $A_t$ is the predicted action at step $t$. Each image is encoded into visual tokens $V_t = \{v_1^t, \dots, v_N^t\}$, and the model conditions on $\{V_1, \dots, V_t\}$ and $\{T_1, \dots, T_t\}$ to generate the next reasoning and action. As $t$ increases, the number of visual tokens grows linearly, introducing substantial redundancy due to high similarity between consecutive screenshots. As illustrated in Figure~\ref{fig:method}, our goal is to construct a reduced set of tokens $V_t' \subseteq V_t$ that preserves task-relevant information while removing redundancy. To achieve this, for each step $t$, we compute a binary mask $\mathbf{m}_t \in \{0,1\}^N$ by comparing two corresponding patches from $I_{t-1}$ and $I_t$, where $\mathbf{m}_t[j]=1$ indicates that the $j$-th patch in $I_t$ is retained. The filtered tokens are then given by $V_t' = V_t[\mathbf{m}_t]$, and the model uses $\{V_1', \dots, V_t'\}$ together with $\{T_1, \dots, T_t\}$ to generate the next action and reasoning.

\begin{algorithm*}[t]
\caption{ReVision Visual Token Dropping}
\label{alg:revision_inference}
\small
\begin{algorithmic}[1]
\Require \textbf{Input:} current step $N$, image window size $k$, task information $q$, previous actions $\mathcal{A}_{1:N-1}$, previous reasoning $\mathcal{R}_{1:N-1}$, recent images $\mathcal{I}_{N-k+1:N} = [I_{N-k+1}, \dots, I_N]$

\State $\mathbf{x} \gets \textproc{Tokenizer}\!\left(q, \mathcal{A}_{1:N-1}, \mathcal{R}_{1:N-1}, \langle\text{image}\rangle^k\right)$
\Comment{Insert one \textsc{<image>} placeholder per image in its actual step}

\State $\mathcal{V} \gets [\ ]$, \quad $\mathcal{P} \gets [\ ]$

\For{$i = N-k+1$ to $N$}
    \State $(\mathbf{u}_i, \mathbf{p}_i) \gets \textproc{VisionEncoder}(I_i)$
    \State $(\mathbf{v}_i, \mathbf{f}_i) \gets \textproc{ProjectionLayer}(\mathbf{u}_i)$
    \Comment{$\mathbf{v}_i$: visual tokens, $\mathbf{f}_i$: patch features, $\mathbf{p}_i$: original position ids}

    \If{$i = N-k+1$}
        \State $\hat{\mathbf{v}}_i \gets \mathbf{v}_i$
        \State $\hat{\mathbf{p}}_i \gets \mathbf{p}_i$
        \Comment{Keep the first image in the window intact}
    \Else
        \State $\mathbf{m}_i \gets \textproc{ReVisionTokenSelection}(\mathbf{f}_{i-1}, \mathbf{f}_i)$
        \Comment{$\mathbf{m}_i$ is a binary indicator of which features/tokens to keep}
        \State $\hat{\mathbf{v}}_i \gets \mathbf{v}_i[\mathbf{m}_i]$
        \State $\hat{\mathbf{p}}_i \gets \mathbf{p}_i[\mathbf{m}_i]$
        \Comment{Retained tokens keep their original position ids}
    \EndIf

    \State append $\hat{\mathbf{v}}_i$ to $\mathcal{V}$
    \State append $\hat{\mathbf{p}}_i$ to $\mathcal{P}$
\EndFor

\State $(\mathbf{e}, \mathbf{pos}) \gets \textproc{BuildMultimodalInput}(\mathbf{x}, \mathcal{V}, \mathcal{P})$
\Comment{Pass all previous actions and reasoning, but only the last $k$ images}
\State \Return $\textproc{LLMDecoder}(\mathbf{e}, \mathbf{pos})$

\end{algorithmic}
\end{algorithm*}

\subsection{Training}

\noindent
\textbf{Training the RTS classifier.}  
We train the ReVision Token Selection (RTS) module as a lightweight three-layer MLP that predicts whether a patch in the current image is redundant given corresponding patch from the previous image. To obtain supervision, we use OmniParserV2~\citep{lu2024omniparserpurevisionbased} to segment screenshots into semantic regions and match regions across consecutive images. This allows us to generate labels based on region overlap (IoU), identifying which patches correspond to unchanged content. We adopt this approach instead of relying on raw pixel or embedding similarity, as region-level matching is more robust to small visual variations (e.g., rendering noise, cursor movement) while still capturing semantic redundancy.

Following the formulation in Section~\ref{sec:problem_formulation} and the procedure outlined in Algorithm~\ref{alg:revision_inference}, we construct training data from AgentNet trajectories by applying RTS to each pair of consecutive images. At each step $t$, RTS compares each patch in $I_t$ with its corresponding patch in $I_{t-1}$ to produce a mask $\mathbf{m}_t$, which is then used to obtain the filtered tokens $V_t'$. The first image in each window is kept unchanged. This results in sequences where the model observes $\{V_1', \dots, V_t'\}$ together with the full textual context $\{T_1, \dots, T_t\}$ to generate the next reasoning and action. We then fine-tune the MLLM on these filtered trajectories, training it to operate under partially observed visual inputs where redundant patches are removed and must be implicitly recovered from previous steps.

\noindent
\textbf{Training MLLM with filtered trajectories.}  
RTS is applied as a plug-in token filtering mechanism on top of MLLMs, following the same formulation and pipeline used during both training and inference (Algorithm~\ref{alg:revision_inference}). At each step $t$, the model conditions on the full textual context $\{T_1, \dots, T_t\}$, while only the most recent $k$ images within a fixed history window are included from the trajectory. Each image $I_t$ is encoded into visual tokens $V_t$, and for consecutive images, RTS compares each patch in $I_t$ with its corresponding patch in $I_{t-1}$ to produce a binary mask $\mathbf{m}_t$, which is used to construct the filtered tokens $V_t' = V_t[\mathbf{m}_t]$. The first image in the window is kept unchanged, while subsequent images retain only non-redundant patches. The model then operates on $\{V_1', \dots, V_t'\}$ together with $\{T_1, \dots, T_t\}$ to generate the next reasoning and action. We construct training samples from AgentNet~\citep{wang2025opencuaopenfoundationscomputeruse} trajectories using this sliding window of $\textit{k}$, ensuring that all trajectories are used while only images within the history window are provided at each step. By training under the same filtered setting used at inference time, the model learns to recover missing visual information from earlier images, enabling efficient use of longer histories without requiring full observations at every step.
\section{Experiments and Results}
\label{sec:experiments}

\subsection{Experimental Settings}
\label{sec:exp_setup}

\begin{figure*}[t]
    \centering
    \includegraphics[width=\textwidth]{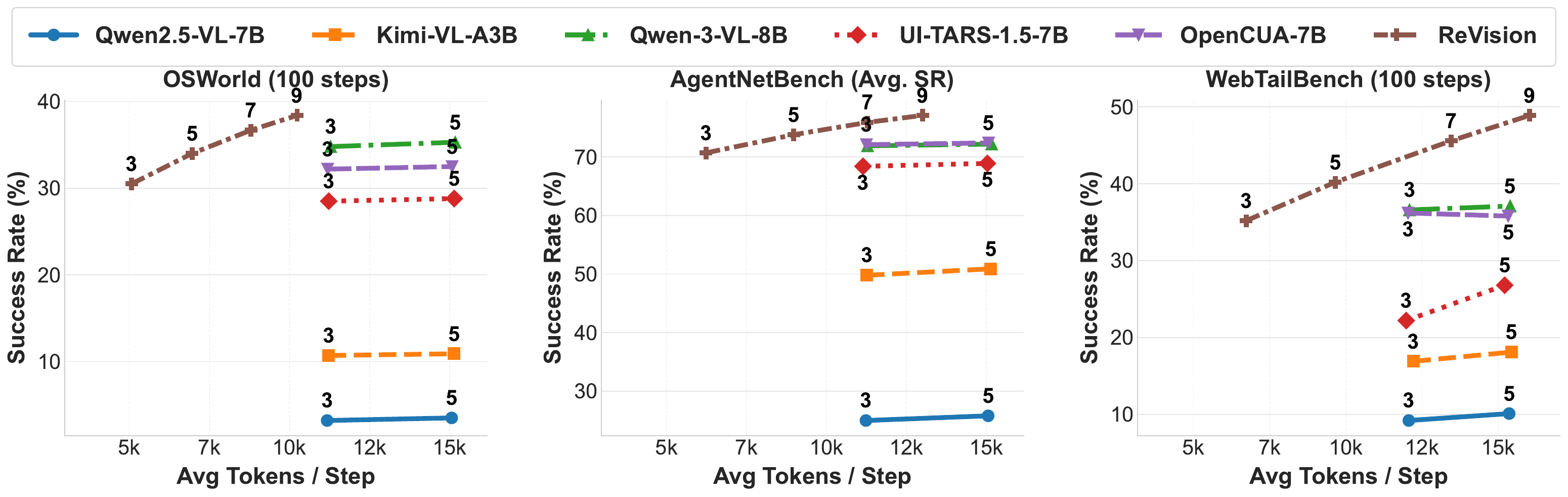}
    \caption{
    Success rate versus average tokens per step across OSWorld at 100 steps, AgentNetBench, and WebTailBench at 100 steps. \name{} consistently achieves high success rates at comparable or lower token budgets, effectively shifting the efficiency frontier. Detailed numerical results are provided in Tables~\ref{tab:osworld_main}, \ref{tab:agentnetbench_main}, and \ref{tab:webtailbench} in Appendix~\ref{app:main_results_tables}. See Figure~\ref{fig:more_token_vs_sr} in Appendix~\ref{app:additional_efficienct_results} for results on OSWorld at 15 steps, 50 steps, and WebTailBench at 50 steps.
    }
    \label{fig:token_vs_sr}
\end{figure*}

\begin{figure*}[t]
    \centering
    \includegraphics[width=\textwidth]{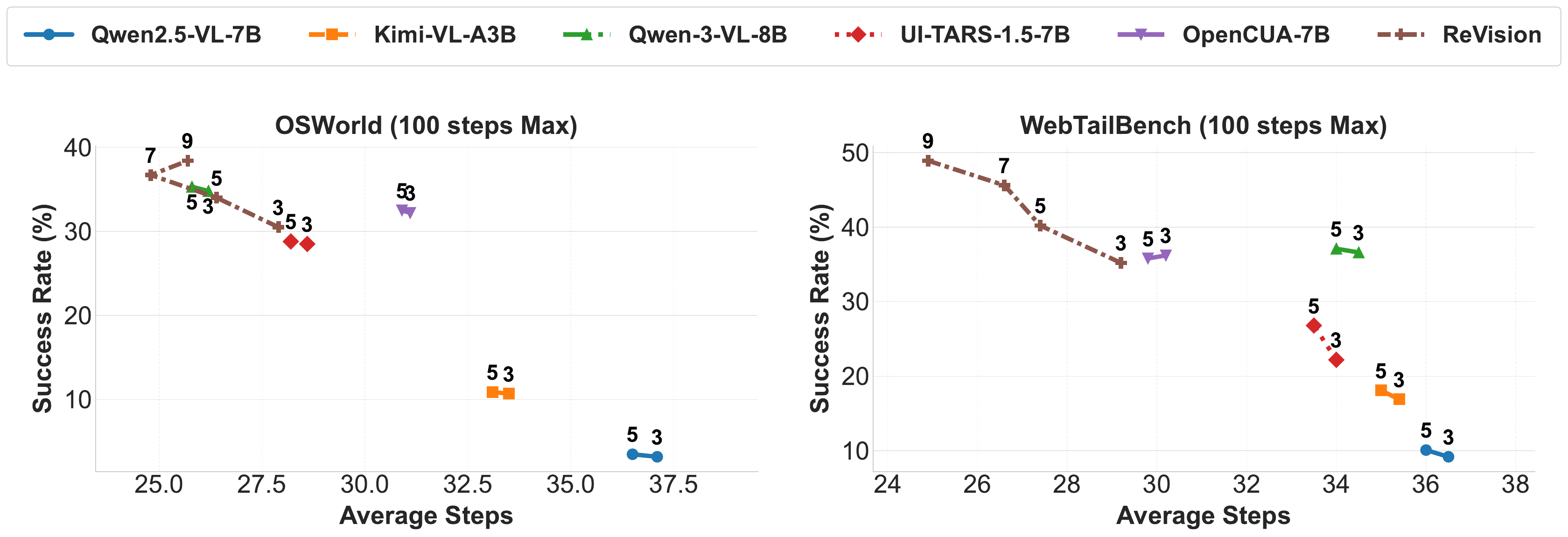}
    \caption{
    Success rate versus average number of steps for OSWorld and WebTailBench at 100 steps. \name{} achieves higher success rates with fewer steps, indicating more efficient decision-making. Detailed numerical results are provided in Tables~\ref{tab:agentnetbench_main} and \ref{tab:webtailbench} in Appendix~\ref{app:main_results_tables}. See Figure~\ref{fig:more_steps_vs_sr} in Appendix~\ref{app:additional_efficienct_results} for results on OSWorld and WebTailBench at 50 steps.
    }
    \label{fig:steps_vs_sr}
\end{figure*}

We build on the OpenCUA framework~\citep{wang2025opencuaopenfoundationscomputeruse} and follow its default setup. We train on AgentNet~\citep{wang2025opencuaopenfoundationscomputeruse}, using a separate model for each history window size $k$ to match training and inference conditions. Training uses standard autoregressive next-token prediction with the same optimizer and hyperparameters as OpenCUA, and decoding temperature is fixed to $T{=}0.0$ to isolate the effect of token filtering. All results are averaged over three runs, with error analysis in Appendix~\ref{app:error_analysis}. Additional training details and metrics are provided in Appendix~\ref{app:training_details}.

\noindent
\textbf{Benchmarks \& Metrics.}
We evaluate on OSWorld~\citep{OSWorld}, AgentNetBench~\citep{wang2025opencuaopenfoundationscomputeruse}, and WebTailBench~\citep{awadallah2025fara}, covering long-horizon desktop and web-based tasks. We report success rate (SR), defined as the percentage of tasks completed successfully. For OSWorld and WebTailBench, we report SR under different step budgets, following their interactive evaluation settings. AgentNetBench is an offline benchmark with fixed trajectories, so we report average SR over its task categories. For WebTailBench, we execute tasks in the OSWorld environment and use an LLM-as-a-judge (\textsc{GPT-4o}~\citep{openai_gpt4o_2024}) to assess step-level correctness and compute SR. See Appendix~\ref{app:benchmark_details} for more details on the benchmarks.

\noindent
\textbf{Baselines.}
We compare \name{} with general vision-language models, including \textsc{Qwen-2.5-VL}~\citep{Qwen2.5-VL}, \textsc{Qwen-3-VL}~\citep{qwen3technicalreport, bai2025qwen3}, and \textsc{Kimi-VL-A3B}~\citep{kimiteam2025kimivltechnicalreport}, as well as specialized UI agents such as \textsc{UI-TARS}~\citep{wang2025ui, qin2025ui} and \textsc{OpenCUA}~\citep{wang2025opencuaopenfoundationscomputeruse}. \textit{For these baselines, we provide all $k$ history images without applying any patch-removal strategy. We do so because these models are trained to process full images, and naive token removal degrades their performance; \textbf{effective token reduction therefore requires model fine-tuning}}. Accordingly, token counts are reported to show that \name{} can achieve strong performance with substantially fewer visual tokens, rather than as a direct efficiency comparison against baselines under token pruning. To isolate the effect of token removal, we include \textbf{\name{} No Drop}, which uses the same training setup as \name{} but disables patch removal at inference time. All baselines and \name{} variants receive the full history of reasoning traces and actions. Unless otherwise specified, all \name{} models, including \name{} No Drop, use \textsc{Qwen2.5-VL-7B}.

\subsection{Efficiency-Performance Trade-offs}
\label{subsec:efficiency_tradeoff}

We analyze the trade-off between task performance and computational cost through two views: success rate versus token usage and success rate versus trajectory length. At each step, the agent receives the last $k$ images along with the full textual context (reasoning and actions), where the history constraint applies only to images. 

\noindent
\textbf{Performance vs. Token Usage.}
Figure~\ref{fig:token_vs_sr} shows that adding more history images yields only marginal SR gains for baselines while substantially increasing token usage, suggesting that redundant information accumulates across consecutive GUI observations. \name{} scales differently: it uses far fewer tokens while consistently matching or outperforming baselines across all benchmarks. It improves SR by up to \textbf{7 points} on OSWorld and AgentNetBench and up to \textbf{14 points} on WebTailBench, while using \textbf{34\% fewer tokens per image}. This efficiency enables \textbf{9-image histories} under a token budget comparable to \textbf{5-image} baselines. On WebTailBench, \name{} reaches nearly \textbf{50\% SR} (\textbf{42\% relative improvement}) versus below \textbf{30\%} for strong baselines. These results show that many visual tokens in GUI trajectories are redundant, and that removing them can improve—not hurt—performance by producing a more compact and informative history. See Appendix~\ref{app:case_study} for a qualitative case study.

\begin{figure*}[t]
    \centering
    \includegraphics[width=\textwidth]{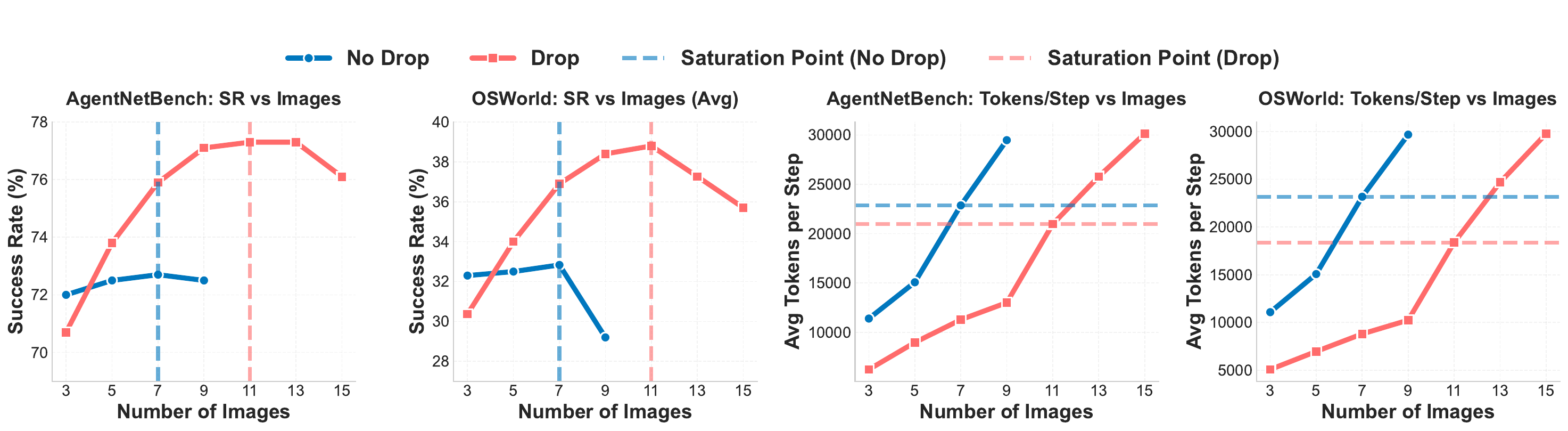}
    \caption{\textbf{Visual History Scaling.} As the number of history images increases, the No Drop baseline saturates early due to rising token usage, while \name{} removes redundant tokens, delaying saturation and achieving higher performance under a similar budget.}
    \label{fig:saturation}
    \vspace{-1em}
\end{figure*}

\noindent
\textbf{Performance vs. Trajectory Length.}
Figure~\ref{fig:steps_vs_sr} shows the relationship between success rate and trajectory length, measuring how efficiently models solve tasks as history is incorporated. \name{} consistently achieves higher success rates with fewer steps. On OSWorld, \name{} reduces average trajectory length by up to \textbf{4 steps} while improving success rate. On WebTailBench, the gains are stronger: \name{} reduces trajectories by up to \textbf{4 steps} and improves success rate by up to \textbf{14 points}. While strong baselines require 33--37 steps and still remain below \textbf{40\%} SR, \name{} reaches nearly \textbf{50\%} SR with only \textbf{$\sim$25--30 steps}. One exception is OSWorld with \textbf{9-image histories}, where \name{} shows a slight increase in steps despite strong performance, possibly due to over-reasoning from longer histories. Since this pattern does not appear at lower step budgets (SR@15 and SR@50) or on other benchmarks, it appears specific to longer-horizon OSWorld settings. Overall, these results show that redundancy-aware token filtering improves both task success and decision efficiency.

\subsection{Effect of Using Different Token Selection}
\label{subsec:different_token_selection}

In Table~\ref{tab:token_selection_main}, we compare \name{} with random, spiral, pixel-similarity, and embedding-similarity (\textsc{Qwen2.5-VL-7B} and \textsc{DinoV2-base}~\cite{oquab2024dinov2learningrobustvisual}) token selection strategies. Naive dropping reduces tokens but consistently hurts performance, with aggressive removal (Random 90\%) causing catastrophic failure. Pixel-based filtering compresses more but relies on noisy low-level similarity, while embedding-based methods better preserve performance but still do not outperform no dropping. In contrast, \textbf{RTS provides the best performance--efficiency trade-off}, improving success rate over no dropping ($73.8$ vs.\ $72.5$ on AgentNet; $34.0$ vs.\ $32.3$ on OSWorld) while reducing tokens per step by \textbf{48\%} on average. OmniParserV2 achieves the highest success rates but incurs \textbf{high latency} ($>550$ ms), whereas \name{} obtains comparable gains with only $\sim$22 ms latency. These results highlight the importance of \emph{semantic and temporal awareness} for effective token selection (see Appendix~\ref{app:omniparser_ablation}).

\begin{table}[h]
    \centering
    \renewcommand{\arraystretch}{1.1}
    \resizebox{\linewidth}{!}{%
    \begin{tabular}{l|cc|cc|c}
        \hline
        \multirow{2}{*}{\textbf{Strategy}} 
        & \multicolumn{2}{c|}{\textbf{AgentNet}} 
        & \multicolumn{2}{c|}{\textbf{OSWorld}} 
        & \multirow{2}{*}{\textbf{Lat. (ms)}} \\ \cline{2-5} 
        & \textbf{SR} & \textbf{Tok/Step}
        & \textbf{SR@100} & \textbf{Tok/Step}
        & \\ 
        \hline

        \rowcolor{gray!20}
        No Drop 
        & 72.5 & 15076
        & 32.3 & 15071
        & 0 \\ \hline

        Random (50\%) 
        & 67.9 & 9952
        & 27.8 & 9788
        & 0 \\

        \textbf{Random (90\%)} 
        & 18.9 & 4234
        & 4.6 & 4385
        & 0 \\

        Spiral (50\%) 
        & 69.4 & 9821
        & 29.0 & 9662
        & 0 \\ \hline

        Pixel 
        & 68.4 & 8213
        & 28.6 & 6125
        & 20 \\ \hline

        Qwen2.5-VL-7B (CosSim) 
        & 72.3 & 9424
        & 32.1 & 7624
        & 7 \\

        DINOv2-base (CosSim) 
        & 71.7 & 9682
        & 31.4 & 7915
        & 28.5 \\

        \rowcolor{cyan!4}
        RTS + OmniParserV2 
        & 74.6 & 8420
        & 35.2 & 6485
        & 565 \\

        \rowcolor{cyan!12}
        \textbf{RTS (Ours)} 
        & 73.8 & 8975
        & 34.0 & 6963
        & 22.5 \\

        \hline
    \end{tabular}
    }
    \caption{\textbf{Comparison of different token selection strategies.} Lat. is average latency. Moderate dropping reduces tokens but degrades performance, while aggressive removal (Random 90\%) causes catastrophic failure. RTS achieves the best performance-efficiency trade-off with low average latency, whereas region-based methods (OmniParserV2) improve performance at significantly higher cost. See Appendix~\ref{app:token_selection_qualitative} for qualitative analysis on different removal strategies.}
    \label{tab:token_selection_main}
\end{table}

\subsection{Visual History Scaling and Saturation}
\label{subsec:history_saturation}

We analyze how increasing the number of history images affects performance and token usage for the \textit{No Drop} baseline and \name{}. As shown in Figure~\ref{fig:saturation}, performance initially improves with longer histories but eventually saturates: the \textit{No Drop} baseline peaks earlier, around 7 images, and then declines, while \name{} continues improving up to larger windows, around 11 images, before plateauing. Notably, saturation aligns more closely with total context length than with the number of images, occurring at approximately \textbf{23k tokens} across benchmarks. By removing redundant visual tokens, \name{} compresses the context and delays saturation, enabling more effective use of longer histories within the same token budget.

To better understand this behavior, we compare two token-removal directions in Table~\ref{tab:forward_backward}. The default \name{} uses \textit{forward} removal, which removes redundant patches from later screenshots, including the final observation. We also train a \textit{backward} variant that keeps the final image intact and removes tokens from earlier history images. Forward removal performs best for $H=5$ and $H=9$ across all benchmarks, while backward removal degrades performance, especially with longer histories. This suggests that an overly complete final image can reduce the model's reliance on history; filtering it encourages the use of temporally distributed evidence. These results support our design choice and show that \name{} improves long-horizon reasoning by both reducing tokens and promoting better use of visual history. See Appendix~\ref{app:window_generalizartion} for window-size generalization analysis.

\begin{table}[t]
\centering
\small
\setlength{\tabcolsep}{6pt}
\renewcommand{\arraystretch}{1.1}
\resizebox{\linewidth}{!}{%
\begin{tabular}{llccc}
\toprule
\textbf{Benchmark} & \textbf{History} & \textbf{NoDrop} & \textbf{Forward} & \textbf{Backward} \\
\midrule
\multirow{3}{*}{AgentNetBench}
& $H=3$ & \textbf{72.0} & 70.7 & 71.4 \\
& $H=5$ & 72.5 & \textbf{73.8} & 70.2 \\
& $H=9$ & 72.5 & \textbf{77.1} & 68.9 \\
\midrule
\multirow{3}{*}{WebTailBench}
& $H=3$ & \textbf{36.3} & 35.2 & 35.6 \\
& $H=5$ & 36.0 & \textbf{40.2} & 33.8 \\
& $H=9$ & 35.1 & \textbf{48.9} & 31.6 \\
\midrule
\multirow{3}{*}{OSWorld}
& $H=3$ & \textbf{32.2} & 30.5 & 31.4 \\
& $H=5$ & 32.3 & \textbf{34.0} & 29.6 \\
& $H=9$ & 29.4 & \textbf{38.4} & 27.8 \\
\bottomrule
\end{tabular}%
}
\caption{Effect of forward vs.\ backward token selection across benchmarks and history lengths. Best results within each benchmark and history setting are bolded.}
\label{tab:forward_backward}
\vspace{-1em}
\end{table}
\subsection{Ablations}
\label{sec:ablations}

\noindent
\textbf{\name{} Generalizes Across Models.}
We evaluate whether \textbf{\name{}} generalizes beyond a single backbone by comparing its performance across two model families: \textit{Qwen2.5-VL-7B} and \textit{Qwen3-VL-8B}. We report results for history window sizes of 3 and 5 images on \textit{OSWorld} and \textit{AgentNetBench}. As shown in Table~\ref{tab:generalization_main}, increasing the history size consistently improves performance for both model families while maintaining predictable scaling in token usage. Notably, the relative gains are consistent across benchmarks and architectures, indicating that \name{} generalizes effectively beyond a specific backbone. We observe that the improvement margins for \textit{Qwen3-VL-8B} are slightly smaller, which may be attributed to its stronger baseline performance on computer-use tasks, leaving less room for improvement.

\begin{table}[h]
    \centering
    \setlength{\tabcolsep}{6pt}
    \resizebox{\linewidth}{!}{%
        \begin{tabular}{l|c|cc}
        \toprule
        \textbf{Base Model} & \textbf{History} 
        & \textbf{OSWorld} 
        & \textbf{AgentNetBench} \\
        \midrule

        \multirow{2}{*}{Qwen2.5-VL-7B}
        & 3 & 30.5 & 70.7 \\
        & 5 & 34.0 & 73.8 \\

        \midrule

        \multirow{2}{*}{Qwen3-VL-8B}
        & 3 & 34.1 & 73.5 \\
        & 5 & 36.7 & 76.0 \\

        \bottomrule
        \end{tabular}
    }
    \caption{Generalization of \name{} across model families using 3 and 5 image history windows. We report SR@100 for OSWorld and average SR for AgentNetBench. Full results are provided in Appendix~\ref{app:generalization_model}.}
    \label{tab:generalization_main}
\end{table}

\begin{table}[h]
    \centering
    \setlength{\tabcolsep}{6pt}
    \resizebox{\linewidth}{!}{%
    \begin{tabular}{lccc}
        \toprule
        \textbf{Model} & \textbf{OSWorld-G} & \textbf{ScreenSpot-Pro} & \textbf{UI-Vision} \\
        \midrule
        Qwen2.5-VL-7B & 31.3 & 27.8 & 0.85 \\
        Qwen3-VL-7B   & 57.8 & 55.3 & 27.6 \\
        \midrule
        \name{} (Qwen2.5-VL-7B) & 31.1 & 27.6 & 0.83 \\
        \name{} (Qwen3-VL-8B)   & 57.5 & 55.6 & 27.2 \\
        \bottomrule
    \end{tabular}
    }
    \caption{Performance comparison in the single-image setting across GUI grounding benchmarks. \name{} achieves comparable performance to the base models, confirming that training with temporal token filtering does not degrade single-image grounding ability.}
    \label{tab:single_image_eval}
\end{table}

\noindent
\textbf{\name{} Does not Hurt Performance with a Single Image.}
To verify that training with \textbf{\name{}} does not degrade performance in the standard single-image setting, we evaluate our models on four GUI grounding benchmarks: \textit{OSWorld-G}~\citep{xie2025scalingcomputerusegroundinguser}, \textit{ScreenSpot-Pro}~\citep{li2025screenspotpro}, and \textit{UI-Vision}~\citep{nayak2025uivisiondesktopcentricguibenchmark}. In this setting, no historical images are provided, and thus \name{}’s token filtering mechanism is effectively inactive at inference time. This experiment isolates whether the modified training distribution, where redundant visual tokens are removed across trajectories, negatively impacts performance when only a single screenshot is available. As shown in Table~\ref{tab:single_image_eval}, \name{} achieves performance comparable to the base models across all benchmarks, with only minor variations. These results indicate that training with filtered visual inputs does not harm the model’s grounding ability in the single-image regime.
\section{Conclusion}
\label{sec:conclusion}

In this work, we introduced \textbf{ReVision}, a redundancy-aware history representation for computer-use agents that reduces unnecessary visual tokens by explicitly modeling temporal redundancy across consecutive screenshots. Our results show that a substantial portion of visual context in GUI trajectories is redundant, and that removing these tokens improves both efficiency and task performance. Across multiple benchmarks, ReVision consistently achieves higher success rates while using fewer tokens and shorter trajectories, demonstrating that better history representation can improve decision-making in long-horizon computer-use tasks. More broadly, our findings suggest that the key challenge in scaling visual reasoning is not simply the number of past images, but how much useful information can be preserved within a limited context budget. Looking forward, an important direction is to extend redundancy modeling beyond time to also capture spatial redundancy within screenshots, and to better understand the mechanisms behind performance saturation in long-context multimodal reasoning.


\section*{Limitations}
\name{} has a few limitations worth noting. First, its effectiveness depends on the quality of the learned redundancy predictor. Although RTS is lightweight and trained to identify temporally redundant visual patches, errors in this classifier can remove patches that later become important for GUI reasoning, especially when small visual changes encode critical state transitions. This risk is particularly relevant for interfaces with subtle updates, dense text, small icons, or dynamically changing elements. Second, \name{} is designed around temporal redundancy in screenshot histories, and therefore provides the largest gains in long-horizon GUI tasks where consecutive screenshots contain substantial visual overlap. In settings with highly dynamic visual environments, rapid scene changes, or tasks that require detailed comparison of every frame, the achievable compression may be lower. Similarly, while preserving positional indices helps maintain compatibility with m-RoPE and spatial reasoning, token removal is still a lossy operation and may affect tasks requiring pixel-level precision. Finally, our experiments focus on GUI-based computer-use agents and vision-language models. While the core idea of removing temporally redundant visual tokens may generalize to other multimodal settings, such as video understanding or robotic perception, we do not claim that the current implementation directly transfers to those domains without additional adaptation.

\section*{Ethical Considerations}
\label{sec:ethics}

\paragraph{Potential Risks.}
\name{} is designed to improve the efficiency of multimodal computer-use agents by reducing redundant visual tokens from screenshot histories. More efficient agents can make long-horizon GUI automation cheaper and more accessible, but they may also amplify risks associated with autonomous computer-use systems. For example, if deployed without appropriate safeguards, such agents could perform unintended actions in user interfaces, mishandle sensitive information shown on screen, or interact with web services in ways that violate user expectations or platform policies. Since \name{} changes the visual context available to the agent, incorrect token pruning could also lead to mistaken actions when important interface details are removed. We therefore recommend that \name{} be used with human oversight, action confirmation mechanisms for sensitive operations, and careful evaluation before deployment in real-world environments.

\paragraph{Privacy and Data Considerations.}
\name{} operates on screenshot trajectories, which may contain sensitive information such as names, emails, account details, documents, or other private user content. Our work uses publicly available benchmarks and does not collect new private user data. However, future deployments should treat screenshots as sensitive data, avoid unnecessary storage of raw trajectories, and apply appropriate anonymization, access control, and data retention policies. Efficiency improvements should not be interpreted as eliminating the need for privacy-preserving handling of visual interaction data.

\paragraph{Use of AI Assistance.}
During the preparation of this manuscript, we used large language models for paraphrasing and improving the writing quality of the paper, and for assisting with figure and caption refinement. All ideas, technical contributions, experimental design, implementation, and conclusions are entirely original and the work of the authors.

\paragraph{Reproducibility Statement.}
We describe the architecture of \name{}, the RTS training setup, token selection procedure, benchmarks, baselines, and evaluation settings in Section~\ref{sec:exp_setup}. We release our code to support reproducibility. All external benchmarks, datasets, models, and baseline artifacts used in this paper are publicly available, and our use of them is limited to research evaluation, consistent with their original licenses, terms of use, and intended access conditions. Our released artifacts are intended for research purposes and for reproducing the experiments reported in this paper.

\paragraph{Use of Existing Artifacts and Data.}
We use only publicly available benchmarks, datasets, models, and baseline artifacts. We do not collect new user data or attempt to identify individuals. Some benchmark trajectories may contain web-derived or interface-specific text from the original datasets; we follow the original preprocessing, licenses, and access conditions of those datasets. All benchmarks, datasets, and baselines used in this paper are openly available, and our use of them is consistent with research evaluation and reproducibility.



\bibliography{custom}



\appendix

\section{Benchmark Details}
\label{app:benchmark_details}

We evaluate \name{} on three computer-use benchmarks: OSWorld, WebTailBench, and AgentNetBench. These benchmarks cover complementary settings, including interactive desktop control, web-based long-horizon tasks, and offline trajectory-based evaluation. Table~\ref{tab:benchmark_stats} summarizes their main characteristics.

\begin{table*}[t]
\centering
\small
\setlength{\tabcolsep}{5pt}
\renewcommand{\arraystretch}{1.12}
\resizebox{\linewidth}{!}{%
\begin{tabular}{lcccc}
\toprule
\textbf{Benchmark} 
& \textbf{Setting} 
& \textbf{\# Tasks} 
& \textbf{Evaluation} 
& \textbf{Reported Metric} \\
\midrule
OSWorld 
& Desktop GUI 
& 369 
& Interactive execution 
& SR@15 / SR@50 / SR@100 \\
WebTailBench 
& Web GUI 
& 609 
& Interactive execution 
& SR@50 / SR@100 \\
AgentNetBench 
& Desktop GUI 
& 1091 
& Offline trajectory evaluation 
& Func. SR / Content SR / Coord. SR / Avg. SR \\
\bottomrule
\end{tabular}%
}
\caption{Summary of the benchmarks used in our evaluation. OSWorld and WebTailBench are evaluated interactively under different step budgets, while AgentNetBench is an offline benchmark with fixed trajectories. SR denotes success rate.}
\label{tab:benchmark_stats}
\end{table*}

\paragraph{OSWorld}~\citep{OSWorld} evaluates agents in realistic desktop environments with tasks involving web browsing, file manipulation, office applications, system settings, and multi-application workflows. Each task provides an initial environment state and an instruction, and the agent must interact with the GUI until the task is completed or the step budget is exhausted. We report success rate under multiple step budgets, SR@15, SR@50, and SR@100, to measure both short-horizon and longer-horizon task completion. This benchmark is particularly useful for evaluating \name{} because many tasks require reasoning over repeated screenshots where large portions of the interface remain unchanged across steps.

\paragraph{WebTailBench}~\citep{awadallah2025fara} focuses on web-based computer-use tasks and contains longer-tailed task categories that are underrepresented in existing GUI-agent benchmarks. We run WebTailBench in the OSWorld-style interactive environment and report SR@50 and SR@100. Since WebTailBench tasks often require navigating multi-step web workflows, they benefit from maintaining useful visual history. At the same time, consecutive browser screenshots contain substantial redundancy, making the benchmark a strong testbed for evaluating whether \name{} can preserve temporally useful evidence while reducing unnecessary visual tokens.

\paragraph{AgentNetBench}~\citep{wang2025opencuaopenfoundationscomputeruse} is an offline benchmark derived from held-out AgentNet trajectories. Unlike OSWorld and WebTailBench, it does not require live environment interaction during evaluation; instead, models are evaluated on fixed trajectory states and asked to predict appropriate actions. This setting enables faster and more reproducible evaluation across model variants. We report the average success rate across task categories. AgentNetBench complements the interactive benchmarks by isolating the model's ability to reason over visual histories without confounding factors from environment execution.
\section{Detailed Results Tables}
\label{app:main_results_tables}

We provide the full numerical results corresponding to the efficiency--performance trade-off analysis in Section~\ref{subsec:efficiency_tradeoff}.

\paragraph{OSWorld.}

Table~\ref{tab:osworld_main} reports detailed results on OSWorld across general VLMs, UI agents, and ReVision. Across all model families, increasing the number of history images leads to consistent but diminishing improvements in success rate. Moving from one to three images provides the largest gains, while increasing to five images yields only marginal improvements despite a substantial increase in token usage (e.g., from $\sim$4k to $\sim$15k tokens per step). This trend suggests that short-term visual context is beneficial, but additional history quickly becomes redundant. A similar pattern is observed in trajectory length, where increasing history results in only minor reductions in the number of steps, indicating limited improvements in action efficiency.

ReVision exhibits a different scaling behavior. Without token dropping, it matches the performance of the corresponding baselines, confirming that the training procedure does not degrade performance. When redundant visual tokens are removed, ReVision consistently improves both efficiency and performance. For example, with Qwen2.5-VL-7B and a 5-image history, ReVision improves 50-step success rate from 34.5 to 35.9 while reducing tokens per step by more than 2$\times$ (15,071 $\rightarrow$ 6,963) and decreasing the average number of steps (22.7 $\rightarrow$ 19.8). Similar trends are observed across step budgets and backbones. These results indicate that a significant portion of visual tokens is redundant, and that removing them not only reduces computational cost but also leads to more efficient decision-making by enabling the model to better focus on relevant actions and observations.

\begin{table*}[t]
    \centering
    \renewcommand{\arraystretch}{1.15}
    \resizebox{\linewidth}{!}{%
    \begin{tabular}{l|cc|cc|cc|cc|c}
        \multirow{2}{*}{\textbf{Model}}           & \multirow{2}{*}{\textbf{History}} & \multirow{2}{*}{\textbf{Drop}} & \multicolumn{2}{c|}{\textbf{15 Steps}} & \multicolumn{2}{c|}{\textbf{50 Steps}} & \multicolumn{2}{c|}{\textbf{100 Steps}} & \multirow{2}{*}{\textbf{Avg Tokens / Step}} \\ \cline{4-9}
                                                &                                   &                                & \textbf{SR}    & \textbf{Avg Steps}    & \textbf{SR}    & \textbf{Avg Steps}    & \textbf{SR}     & \textbf{Avg Steps}    &                                             \\ \hline
        \rowcolor{gray!20}
        \multicolumn{10}{c}{\textbf{General VLMs}} \\ \hline
        \multirow{3}{*}{Qwen2.5-VL-7B} 
        & 1 & \xmark & 1.6 & 14.5 & 2.2 & 29.6 & 2.7 & 38.2 & 4,013 \\
        & 3 & \xmark & 1.9 & 13.6 & 2.5 & 28.4 & 3.2 & 37.1 & 11,176 \\
        & 5 & \xmark & 2.3 & 13.2 & 2.7 & 28.1 & 3.5 & 36.5 & 15,062 \\
        \multirow{3}{*}{Qwen2.5-VL-32B}           & 1                                 & \xmark          & 2.5            & 13.8                  & 3.3            & 28.5                  & 3.9             & 36.7                  & 4,067                                       \\
                                                & 3                                 & \xmark          & 2.9            & 12.7                  & 3.6            & 27.4                  & 4.2             & 35.6                  & 11,238                                      \\
                                                & 5                                 & \xmark          & 3.0            & 12.5                  & 3.8            & 27.0                  & 4.4             & 35.2                  & 15,087                                      \\
        \multirow{3}{*}{Qwen2.5-VL-72B}           & 1                                 & \xmark          & 3.6            & 13.4                  & 4.4            & 28.1                  & 5.0             & 36.2                  & 4,109                                       \\
                                                & 3                                 & \xmark          & 3.9            & 12.6                  & 4.7            & 27.3                  & 5.3             & 35.4                  & 11,307                                      \\
                                                & 5                                 & \xmark          & 4.1            & 12.4                  & 4.9            & 27.0                  & 5.5             & 35.0                  & 15,183                                      \\
        \multirow{3}{*}{Kimi-VL-A3B}              & 1                                 & \xmark          & 8.5            & 12.6                  & 9.7            & 26.5                  & 10.3            & 34.1                  & 4,094                                       \\
                                                & 3                                 & \xmark          & 9.2            & 11.8                  & 10.0           & 25.9                  & 10.7            & 33.5                  & 11,221                                      \\
                                                & 5                                 & \xmark          & 9.5            & 11.6                  & 10.2           & 25.6                  & 10.9            & 33.1                  & 15,136                                      \\
        \multirow{3}{*}{Qwen-3-VL-8B} 
        & 1 & \xmark & 29.8 & 12.8 & 34.2 & 20.7 & 33.9 & 27.3 & 4,061 \\
        & 3 & \xmark & 30.2 & 11.9 & 35.1 & 19.8 & 34.8 & 26.2 & 11,284 \\
        & 5 & \xmark & 30.9 & 11.6 & 35.6 & 19.2 & 35.3 & 25.8 & 15,171 \\
        \multirow{3}{*}{Qwen-3-VL-32B}
        & 1 & \xmark & 33.2 & 10.4 & 41.1 & 21.4 & 41.0 & 28.4 & 4,133 \\
        & 3 & \xmark & 39.8 & 9.3  & 42.7 & 19.6 & 42.4 & 27.8 & 11,346 \\
        & 5 & \xmark & 40.3 & 9.1  & 42.8 & 18.9 & 42.5 & 27.3 & 15,229 \\
        \multirow{3}{*}{Qwen3-VL-30B-A3B}
        & 1 & \xmark & 27.9 & 10.6 & 32.4 & 23.6 & 31.0 & 30.8 & 4,117 \\
        & 3 & \xmark & 35.2 & 9.5  & 39.8 & 22.6 & 38.6 & 30.0 & 11,318 \\
        & 5 & \xmark & 35.8 & 9.3  & 40.4 & 22.2 & 39.2 & 29.6 & 15,198 \\ \hline
        \rowcolor{green!12}
        \multicolumn{10}{c}{\textbf{UI Agents}} \\ \hline
        \multirow{3}{*}{OpenCUA}                  & 1                                 & \xmark          & 23.8           & 11.2                  & 27.4           & 24.6                  & 26.0            & 31.8                  & 3,994                                       \\
                                                & 3                                 & \xmark          & 30.5           & 10.1                  & 34.1           & 23.8                  & 32.2            & 31.1                  & 11,193                                      \\
                                                & 5                                 & \xmark          & 30.8           & 10.0                  & 34.4           & 23.6                  & 32.5            & 30.9                  & 15,075                                      \\
        \multirow{3}{*}{UI-TARS-72B-DPO}
        & 1 & \xmark & 20.7 & 12.2 & 23.7 & 25.6 & 23.5 & 32.9 & 3,967 \\
        & 3 & \xmark & 20.9 & 11.7 & 24.4 & 25.0 & 24.2 & 32.6 & 11,159 \\
        & 5 & \xmark & 21.1 & 11.4 & 24.9 & 24.6 & 24.6 & 31.4 & 15,032 \\ 
        \multirow{3}{*}{UI-TARS-1.5-7B}
        & 1 & \xmark & 24.1 & 11.4 & 27.3 & 25.2 & 27.1 & 29.1 & 4,058 \\
        & 3 & \xmark & 25.5 & 10.9 & 28.5 & 24.6 & 28.5 & 28.6 & 11,227 \\
        & 5 & \xmark & 26.0 & 10.7 & 29.3 & 24.3 & 28.8 & 28.2 & 15,141 \\\hline
        \rowcolor{cyan!12}
        \multicolumn{10}{c}{\textbf{ReVision}} \\ \hline
        \multirow{4}{*}{ReVision} & 3                                 & \xmark          & 30.6           & 9.8                   & 34.1           & 22.9                  & 32.2            & 29.8                  & 11,078                                      \\
                                                & 3                                 & \cmark          & 28.6           & 8.9                   & 32.0           & 21.0                  & 30.5            & 27.9                  & 5,074                                       \\
                                                & 5                                 & \xmark          & 30.7           & 9.7                   & 34.5           & 22.7                  & 32.3            & 29.7                  & 15,071                                      \\
                                                & 5                                 & \cmark          & 32.1           & 8.3                   & 35.9           & 19.8                  & 34.0            & 26.4                  & 6,963                                       \\
                                
    \end{tabular}
    }
    \caption{OSWorld results across general VLMs, UI agents, and ReVision. For ReVision, we only report settings where the training window size matches the number of history images. A checkmark indicates redundant history token dropping, while a cross indicates no dropping.}
    \label{tab:osworld_main}
\end{table*}

\paragraph{AgentNetBench.}
Table~\ref{tab:agentnetbench_main} reports results on AgentNetBench across general VLMs, UI agents, and ReVision. Similar to OSWorld, increasing the number of history images leads to consistent improvements across all models, with gains saturating beyond 3 to 5 images. This trend holds across coordinate, content, and functional success rates, indicating that additional short-term visual context is beneficial but provides diminishing returns as history grows. Compared to OSWorld, the improvements from longer history are more stable and less sensitive, reflecting the offline nature of AgentNetBench where trajectories are fixed and less prone to compounding errors.

ReVision again exhibits a different scaling behavior. Without token dropping, it matches the corresponding baselines, confirming that the training setup does not degrade performance. When redundant visual tokens are removed, ReVision consistently improves performance across all metrics. For example, with Qwen2.5-VL-7B and a 5-image history, ReVision improves average success rate from 72.5 to 73.8, with gains observed across coordinate, content, and functional metrics. Similar improvements are observed with stronger backbones, such as Qwen-3-VL-8B. These results suggest that removing redundant visual tokens does not harm any specific aspect of the task and instead enables more effective use of the available context. Overall, the improvements are more modest compared to OSWorld, but remain consistent, indicating that redundancy-aware token filtering provides reliable gains even in less challenging, offline evaluation settings.

\begin{table*}[t]
    \centering
    \renewcommand{\arraystretch}{1.15}
    \resizebox{\linewidth}{!}{%
    \begin{tabular}{l|cc|ccc|c}
    \textbf{Model} & \textbf{History} & \textbf{Drop} & \textbf{Coord. SR} & \textbf{Content SR} & \textbf{Func. SR} & \textbf{Avg. SR} \\
    \hline
    \rowcolor{gray!20}
    \multicolumn{7}{c}{\textbf{General VLMs}} \\
    \hline
    Qwen2.5-VL-7B & 1 & \xmark & 25.1 & 18.3 & 11.9 & 23.4 \\
    Qwen2.5-VL-7B & 3 & \xmark & 26.4 & 19.5 & 12.7 & 25.0 \\
    Qwen2.5-VL-7B & 5 & \xmark & 27.0 & 20.1 & 13.3 & 25.8 \\
    Qwen2.5-VL-32B & 1 & \xmark & 38.5 & 28.2 & 18.4 & 35.6 \\
    Qwen2.5-VL-32B & 3 & \xmark & 40.2 & 29.8 & 19.5 & 37.9 \\
    Qwen2.5-VL-32B & 5 & \xmark & 41.0 & 30.5 & 20.2 & 38.7 \\
    Qwen2.5-VL-72B & 1 & \xmark & 42.1 & 30.5 & 20.2 & 38.7 \\
    Qwen2.5-VL-72B & 3 & \xmark & 43.6 & 31.8 & 21.0 & 40.8 \\
    Qwen2.5-VL-72B & 5 & \xmark & 44.4 & 32.4 & 21.7 & 41.5 \\
    Kimi-VL-A3B & 1 & \xmark & 51.6 & 37.8 & 25.9 & 47.2 \\
    Kimi-VL-A3B & 3 & \xmark & 54.0 & 39.5 & 27.2 & 49.8 \\
    Kimi-VL-A3B & 5 & \xmark & 55.2 & 40.4 & 28.1 & 50.9 \\
    Qwen-3-VL-8B & 1 & \xmark & 73.4 & 54.9 & 40.2 & 68.7 \\
    Qwen-3-VL-8B & 3 & \xmark & 74.6 & 58.8 & 43.1 & 71.9 \\
    Qwen-3-VL-8B & 5 & \xmark & 75.2 & 59.3 & 43.4 & 72.2 \\
    Qwen-3-VL-72B & 1 & \xmark & 79.2 & 64.5 & 48.1 & 76.9 \\
    Qwen-3-VL-72B & 3 & \xmark & 82.5 & 67.4 & 50.8 & 79.3 \\
    Qwen-3-VL-72B & 5 & \xmark & 83.6 & 68.3 & 51.9 & 80.2 \\
    Qwen3-VL-30B-A3B & 1 & \xmark & 76.5 & 61.3 & 44.6 & 74.1 \\
    Qwen3-VL-30B-A3B & 3 & \xmark & 80.4 & 65.1 & 48.2 & 77.0 \\
    Qwen3-VL-30B-A3B & 5 & \xmark & 81.3 & 66.0 & 49.1 & 77.8 \\
    \hline
    \rowcolor{green!12}
    \multicolumn{7}{c}{\textbf{UI Agents}} \\
    \hline
    OpenCUA & 1 & \xmark & 72.0 & 55.6 & 39.8 & 68.4 \\
    OpenCUA & 3 & \xmark & 75.8 & 59.4 & 42.5 & 72.1 \\
    OpenCUA & 5 & \xmark & 76.1 & 59.7 & 42.8 & 72.4 \\
    UI-TARS-72B-DPO & 1 & \xmark & 66.2 & 48.1 & 34.7 & 62.3 \\
    UI-TARS-72B-DPO & 3 & \xmark & 67.1 & 49.0 & 35.5 & 63.2 \\
    UI-TARS-72B-DPO & 5 & \xmark & 67.8 & 49.6 & 36.0 & 63.8 \\
    UI-TARS-1.5-7B & 1 & \xmark & 70.4 & 52.3 & 38.8 & 66.9 \\
    UI-TARS-1.5-7B & 3 & \xmark & 71.8 & 53.8 & 40.2 & 68.4 \\
    UI-TARS-1.5-7B & 5 & \xmark & 72.3 & 54.5 & 40.6 & 68.9 \\
    \hline
    \rowcolor{cyan!12}
    \multicolumn{7}{c}{\textbf{ReVision}} \\
    \hline
    ReVision & 3 & \xmark & 75.9 & 59.3 & 42.4 & 72.0 \\
    ReVision & 3 & \cmark & 74.4 & 58.1 & 41.7 & 70.7 \\
    ReVision & 5 & \xmark & 76.2 & 59.8 & 42.9 & 72.5 \\
    ReVision & 5 & \cmark & 77.0 & 61.0 & 44.2 & 73.8 \\
    \end{tabular}%
    }
    \caption{AgentNetBench results across general VLMs, UI agents, and ReVision. For ReVision, we only report settings where the training window size matches the number of history images.}
    \label{tab:agentnetbench_main}
\end{table*}

\paragraph{WebTailBench.}

Table~\ref{tab:webtailbench} reports results on WebTailBench, a long-horizon benchmark designed to evaluate performance on complex multi-step tasks. Compared to OSWorld and AgentNetBench, improvements from increasing history are limited for standard baselines. While moving from one to three images provides moderate gains, further increasing to five images yields marginal or even diminishing returns despite a substantial increase in token usage. For example, OpenCUA improves from 25.8 to 29.5 at 50-step success rate when increasing history from one to three images, but slightly drops to 29.1 at five images while token cost continues to grow significantly. This pattern indicates that longer visual histories introduce redundancy that models struggle to utilize effectively.

ReVision demonstrates a markedly different behavior in this setting. When redundant visual tokens are removed, performance improves consistently as the history length increases, while maintaining significantly lower token usage. For instance, with Qwen2.5-VL-7B, ReVision improves from 28.4 to 40.8 at 50-step success rate as history increases from three to nine images, and from 35.2 to 48.9 at 100 steps. At the same time, it reduces the average number of steps required to complete tasks (e.g., from 27.9 to 23.6 at 5 images), indicating more efficient decision-making. These results show that, unlike baselines, ReVision is able to effectively leverage longer histories by removing redundant visual information. As a result, longer context becomes beneficial rather than detrimental, highlighting that redundancy-aware token filtering is critical for scaling performance in long-horizon computer-use tasks.

\begin{table*}[t]
    \centering
    \renewcommand{\arraystretch}{1.15}
    \resizebox{\linewidth}{!}{%
    \begin{tabular}{l|cc|cc|cc|c}
    \textbf{Model} & \textbf{History} & \textbf{Drop} 
    & \textbf{SR@50} & \textbf{Avg. Steps@50} 
    & \textbf{SR@100} & \textbf{Avg. Steps@100} 
    & \textbf{Avg Tokens / Step} \\
    \hline
    \rowcolor{gray!20}
    \multicolumn{8}{c}{\textbf{General VLMs}} \\
    \hline
    Qwen2.5-VL-7B & 1 & \xmark & 6.5 & 31.8 & 7.8 & 37.2 & 4,213 \\
    Qwen2.5-VL-7B & 3 & \xmark & 7.9 & 31.1 & 9.2 & 36.5 & 12,067 \\
    Qwen2.5-VL-7B & 5 & \xmark & 8.6 & 30.7 & 10.1 & 36.0 & 15,361 \\
    Qwen2.5-VL-32B & 1 & \xmark & 6.2 & 32.1 & 7.1 & 37.4 & 4,287 \\
    Qwen2.5-VL-32B & 3 & \xmark & 6.8 & 31.4 & 7.9 & 36.8 & 12,183 \\
    Qwen2.5-VL-32B & 5 & \xmark & 7.5 & 31.0 & 8.6 & 36.1 & 15,472 \\
    Qwen2.5-VL-72B & 1 & \xmark & 7.5 & 31.7 & 8.6 & 36.9 & 4,349 \\
    Qwen2.5-VL-72B & 3 & \xmark & 8.2 & 31.0 & 9.4 & 36.2 & 12,244 \\
    Qwen2.5-VL-72B & 5 & \xmark & 9.0 & 30.6 & 10.1 & 35.7 & 15,561 \\
    Kimi-VL-A3B & 1 & \xmark & 13.2 & 31.0 & 15.0 & 36.1 & 4,318 \\
    Kimi-VL-A3B & 3 & \xmark & 14.8 & 30.3 & 16.9 & 35.4 & 12,217 \\
    Kimi-VL-A3B & 5 & \xmark & 15.9 & 29.9 & 18.1 & 35.0 & 15,438 \\
    Qwen-3-VL-8B & 1 & \xmark & 26.0 & 26.2 & 32.8 & 35.3 & 4,266 \\
    Qwen-3-VL-8B & 3 & \xmark & 29.8 & 28.7 & 36.6 & 34.5 & 12,089 \\
    Qwen-3-VL-8B & 5 & \xmark & 30.4 & 28.4 & 37.1 & 34.0 & 15,392 \\
    Qwen-3-VL-32B & 1 & \xmark & 34.8 & 27.5 & 41.9 & 34.3 & 4,331 \\
    Qwen-3-VL-32B & 3 & \xmark & 39.6 & 29.6 & 46.8 & 33.4 & 12,214 \\
    Qwen-3-VL-32B & 5 & \xmark & 40.2 & 29.2 & 47.3 & 32.9 & 15,487 \\
    Qwen3-VL-30B-A3B & 1 & \xmark & 32.5 & 27.1 & 39.2 & 34.8 & 4,309 \\
    Qwen3-VL-30B-A3B & 3 & \xmark & 37.4 & 29.1 & 44.3 & 34.0 & 12,173 \\
    Qwen3-VL-30B-A3B & 5 & \xmark & 38.0 & 28.8 & 44.9 & 33.5 & 15,456 \\
    \hline
    \rowcolor{green!12}
    \multicolumn{8}{c}{\textbf{UI Agents}} \\
    \hline
    OpenCUA & 1 & \xmark & 25.8 & 29.7 & 32.4 & 26.4 & 4,237 \\
    OpenCUA & 3 & \xmark & 29.5 & 28.9 & 36.2 & 30.2 & 12,041 \\
    OpenCUA & 5 & \xmark & 29.1 & 28.6 & 35.8 & 29.8 & 15,329 \\
    UI-TARS-72B-DPO & 1 & \xmark & 15.3 & 26.8 & 17.4 & 34.6 & 4,248 \\
    UI-TARS-72B-DPO & 3 & \xmark & 16.9 & 28.4 & 18.2 & 33.7 & 12,107 \\
    UI-TARS-72B-DPO & 5 & \xmark & 17.4 & 28.1 & 19.5 & 33.2 & 15,376 \\
    UI-TARS-1.5-7B-DPO & 1 & \xmark & 16.5 & 26.6 & 20.4 & 35.0 & 4,169 \\
    UI-TARS-1.5-7B-DPO & 3 & \xmark & 18.1 & 28.8 & 22.2 & 34.0 & 11,982 \\
    UI-TARS-1.5-7B-DPO & 5 & \xmark & 18.7 & 28.5 & 26.8 & 33.5 & 15,214 \\
    \hline
    \rowcolor{cyan!12}
    \multicolumn{8}{c}{\textbf{ReVision}} \\
    \hline
    ReVision & 3 & \xmark & 29.6 & 28.1 & 36.3 & 30.3 & 11,907 \\
    ReVision & 3 & \cmark & 28.4 & 25.4 & 35.2 & 29.2 & 6,731 \\
    ReVision & 5 & \xmark & 29.3 & 27.9 & 36.0 & 30.0 & 15,362 \\
    ReVision & 5 & \cmark & 32.8 & 23.6 & 40.2 & 27.4 & 9,651
    \end{tabular}%
    }
    \caption{WebTailBench results on long-horizon tasks. We report success rates at 50 and 100 steps, average trajectory length, and average tokens per step. For ReVision, we report configurations with matched training window and history size, together with the corresponding no-drop controls when available.}
    \label{tab:webtailbench}
\end{table*}
\section{Error Analysis}
\label{app:error_analysis}

To evaluate the stability of our results, we run each model three times and report the standard deviation of success rate in Table~\ref{tab:error_std}. Across all benchmarks and history lengths, the standard deviations remain small, indicating that the observed trends are robust across runs. In particular, \name{} maintains low variance even with longer histories ($H=7$ and $H=9$), where the number of visual tokens and reasoning context increase substantially. This suggests that the gains from redundancy-aware token filtering are not caused by run-specific fluctuations, but reflect a consistent improvement in long-horizon decision-making.

\begin{table*}[h]
\centering
\small
\setlength{\tabcolsep}{5pt}
\renewcommand{\arraystretch}{1.1}
\resizebox{0.7\linewidth}{!}{%
\begin{tabular}{lcccc}
\toprule
\textbf{Model} & \textbf{Hist.} & \textbf{OSWorld} & \textbf{AgentNetBench} & \textbf{WebTailBench} \\
\midrule
\multirow{2}{*}{Qwen2.5-VL-7B}
& $H=3$ & 0.8 & 0.7 & 0.9 \\
& $H=5$ & 0.9 & 0.8 & 1.0 \\
\midrule
\multirow{2}{*}{Kimi-VL-A3B}
& $H=3$ & 1.0 & 0.9 & 1.1 \\
& $H=5$ & 1.1 & 0.8 & 1.2 \\
\midrule
\multirow{2}{*}{Qwen3-VL-8B}
& $H=3$ & 0.7 & 0.6 & 0.8 \\
& $H=5$ & 0.8 & 0.7 & 0.9 \\
\midrule
\multirow{2}{*}{UI-TARS-1.5-7B}
& $H=3$ & 0.9 & 0.8 & 1.0 \\
& $H=5$ & 1.0 & 0.9 & 1.1 \\
\midrule
\multirow{2}{*}{OpenCUA-7B}
& $H=3$ & 0.8 & 0.7 & 0.9 \\
& $H=5$ & 0.9 & 0.8 & 1.0 \\
\midrule
\multirow{4}{*}{\name{}}
& $H=3$ & 0.7 & 0.6 & 0.8 \\
& $H=5$ & 0.8 & 0.6 & 0.9 \\
& $H=7$ & 0.9 & 0.7 & 1.0 \\
& $H=9$ & 1.0 & 0.8 & 1.1 \\
\bottomrule
\end{tabular}%
}
\caption{Standard deviation of success rate over three evaluation runs. We report SR@100 for OSWorld and WebTailBench, and average SR for AgentNetBench. The small standard deviations indicate that the observed trends are stable across runs.}
\label{tab:error_std}
\end{table*}
\section{Generalization Across Different Models}
\label{app:generalization_model}

We provide the full results for ReVision across different history window sizes (3, 5, 7, and 9 images) and model families. These results extend Table~\ref{tab:generalization_main} in the main paper.

\begin{table*}[t]
    \centering
    \small
    \setlength{\tabcolsep}{4pt}
    \renewcommand{\arraystretch}{1.1}
    \resizebox{\textwidth}{!}{%
        \begin{tabular}{l|c|cc|cc|cc}
        \toprule
        \textbf{ReVision Base Model} & \textbf{Hist.} 
        & \multicolumn{2}{c|}{\textbf{WebTailBench}} 
        & \multicolumn{2}{c|}{\textbf{OSWorld}} 
        & \multicolumn{2}{c}{\textbf{AgentNetBench}} \\
        & 
        & \textbf{SR@100} & \textbf{Avg. Tok/Step} 
        & \textbf{SR@100} & \textbf{Avg. Tok/Step} 
        & \textbf{Avg SR} & \textbf{Avg. Tok/Step} \\
        \midrule

        \multirow{4}{*}{Qwen2.5-VL-7B}
        & 3 & 35.2 & 6,731 & 30.5 & 5,074 & 70.7 & 6,235 \\
        & 5 & 40.2 & 9,651 & 34.0 & 6,963 & 73.8 & 8,975 \\
        & 7 & 45.6 & 13,528 & 36.7 & 8,802 & 75.9 & 11,284 \\
        & 9 & 48.9 & 15,283 & 38.4 & 10,241 & 77.1 & 13,012 \\

        \midrule

        \multirow{4}{*}{Qwen3-VL-8B}
        & 3 & 42.1 & 7,209 & 34.1 & 5,396 & 73.5 & 6,654 \\
        & 5 & 46.6 & 10,941 & 36.7 & 7,258 & 76.0 & 9,218 \\
        & 7 & 49.8 & 13,462 & 40.4 & 8,994 & 78.6 & 11,759 \\
        & 9 & 52.4 & 16,031 & 41.6 & 10,723 & 80.2 & 13,921 \\

        \bottomrule
        \end{tabular}
    }
    \caption{Generalization results of ReVision across model families and history sizes.}
    \label{tab:generalization_full}
\end{table*}

\section{Training on a Fixed Context Window Generalizes to Other Window Sizes.} 
\label{app:window_generalizartion}

We analyze whether models trained with \textbf{ReVision} under a fixed visual context window generalize to different context sizes at inference time. Although ReVision is trained with a fixed number of history images, agents in practice may operate under varying context budgets. To study this, we train models with window sizes $w \in {3,5}$ and evaluate them under both matched and mismatched inference windows. Table~\ref{tab:cross_window} reports results on \textit{OSWorld} (100-step success rate) and \textit{AgentNetBench} (average success rate), along with average tokens per step. Performance is best when the training and inference window sizes match, but the drop under mismatched settings is modest. This suggests that ReVision-trained models are robust to changes in the number of visual context images and generalize beyond their training configuration without substantial loss in performance.

\begin{table*}[h]
    \centering
    \small
    \setlength{\tabcolsep}{5pt}
    \resizebox{0.9 \linewidth}{!}{%
        \begin{tabular}{cc cc cc}
        \toprule
        \textbf{Train Window Size} & \textbf{Inference Window Size} 
        & \multicolumn{2}{c}{\textbf{OSWorld}} 
        & \multicolumn{2}{c}{\textbf{AgentNetBench}} \\
        & 
        & \textbf{SR@100} & \textbf{Tok/Step} 
        & \textbf{Avg SR} & \textbf{Tok/Step} \\
        \midrule

        3 & 3 & 30.5 & 5,074 & 70.7 & 6,235 \\
        3 & 5 & 29.1 & 6,755 & 69.7 & 8,437 \\
        5 & 3 & 29.4 & 5,394 & 69.9 & 6,959 \\
        5 & 5 & 34.0 & 6,963 & 73.8 & 8,975 \\

        \bottomrule
    \end{tabular}
    }
    \caption{Cross-window generalization of ReVision on \textbf{Qwen2.5-VL-7B}. Models perform best when training and inference window sizes match, but maintain competitive performance under mismatched settings, indicating robustness to varying visual context sizes.}
    \label{tab:cross_window}
\end{table*}

\section{Qualitative Comparison of Token Selection Strategies}
\label{app:token_selection_qualitative}

\begin{figure*}[t]
    \centering
    \includegraphics[width=\linewidth]{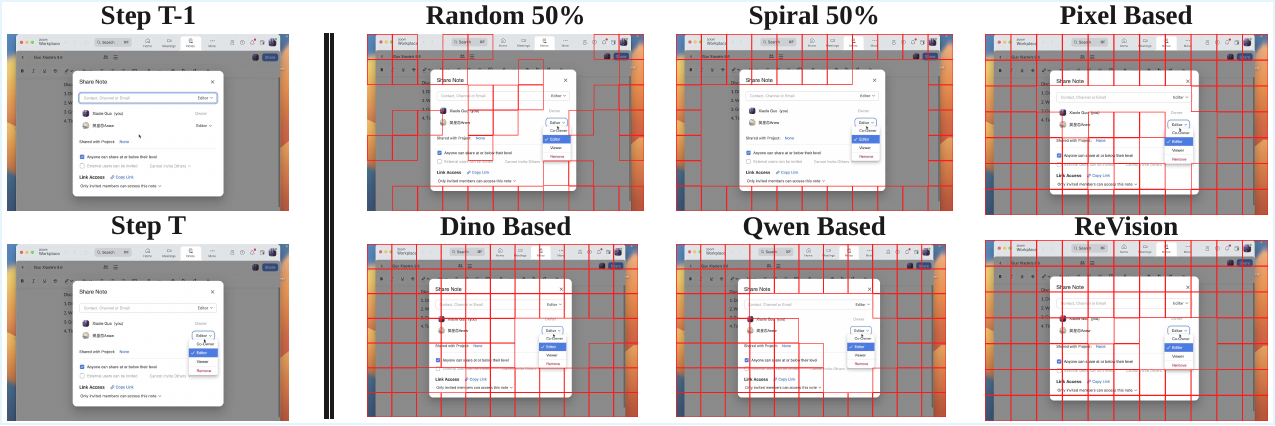}
    \caption{\textbf{Qualitative comparison of token selection strategies.} We show patch retention across different methods for two consecutive steps ($t{-}1$ and $t$). For visualization purposes, we use lower-resolution images, resulting in fewer patches and clearer overlays. Random and spiral strategies remove patches indiscriminately, often discarding important UI elements. Pixel-based similarity removes more patches but fails to preserve fine-grained semantic details. Embedding-based methods (DINO and Qwen) improve consistency but still retain redundant regions. In contrast, ReVision selectively removes temporally redundant patches while preserving task-relevant updates, leading to a more informative and compact visual representation.}
    \label{fig:removal_strategies_appendix}
\end{figure*}

Figure~\ref{fig:removal_strategies_appendix} provides a qualitative comparison of different token selection strategies on a representative trajectory step. We visualize which patches are retained across methods by overlaying the patch grid on the screenshots, where removed regions are suppressed and retained patches are highlighted.

Naive strategies such as \textbf{Random} and \textbf{Spiral} dropping remove patches without considering semantic consistency across time. As a result, they often discard important regions (e.g., UI elements relevant to the task) while retaining redundant background content. This leads to fragmented visual context, where critical information may be missing or partially preserved.

\textbf{Pixel-based similarity} performs more structured filtering by removing patches with low pixel-level changes. However, it is sensitive to small visual variations (e.g., rendering noise, cursor movement), which causes it to either retain redundant regions or remove semantically important details. Consequently, although it achieves stronger token reduction, it often harms downstream reasoning.

\textbf{Embedding-based methods} (DINO and Qwen) provide improved consistency by comparing patch embeddings. These approaches better preserve semantically meaningful regions, but they still struggle to precisely localize task-relevant changes. In particular, they may retain large redundant areas or fail to capture fine-grained updates in the interface.

In contrast, \textbf{ReVision} explicitly models temporal redundancy between corresponding patches across consecutive images. As shown in the figure, it effectively removes unchanged regions while preserving newly updated and task-relevant content. This results in a cleaner and more focused visual representation, where the model can rely on previous images for redundant information while attending to only the necessary updates in the current step.

\section{Additional Efficiency Results}
\label{app:additional_efficienct_results}

\begin{figure*}[t]
    \centering
    \includegraphics[width=\textwidth]{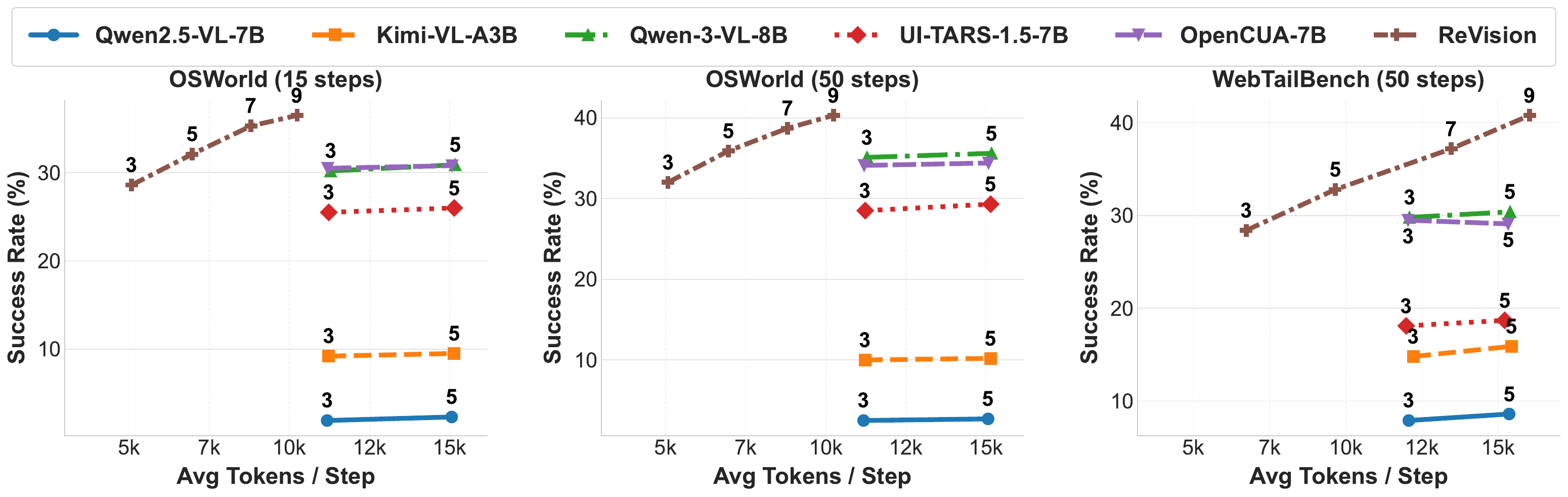}
    \caption{
    Success rate versus average tokens per step across OSWorld at 15 steps, 50 steps, WebTailBench at 50 steps. Detailed numerical results are provided in Tables~\ref{tab:osworld_main}, \ref{tab:agentnetbench_main}, and \ref{tab:webtailbench} in Appendix~\ref{app:main_results_tables}.
    }
    \label{fig:more_token_vs_sr}
\end{figure*}

\begin{figure*}[t]
    \centering
    \includegraphics[width=\textwidth]{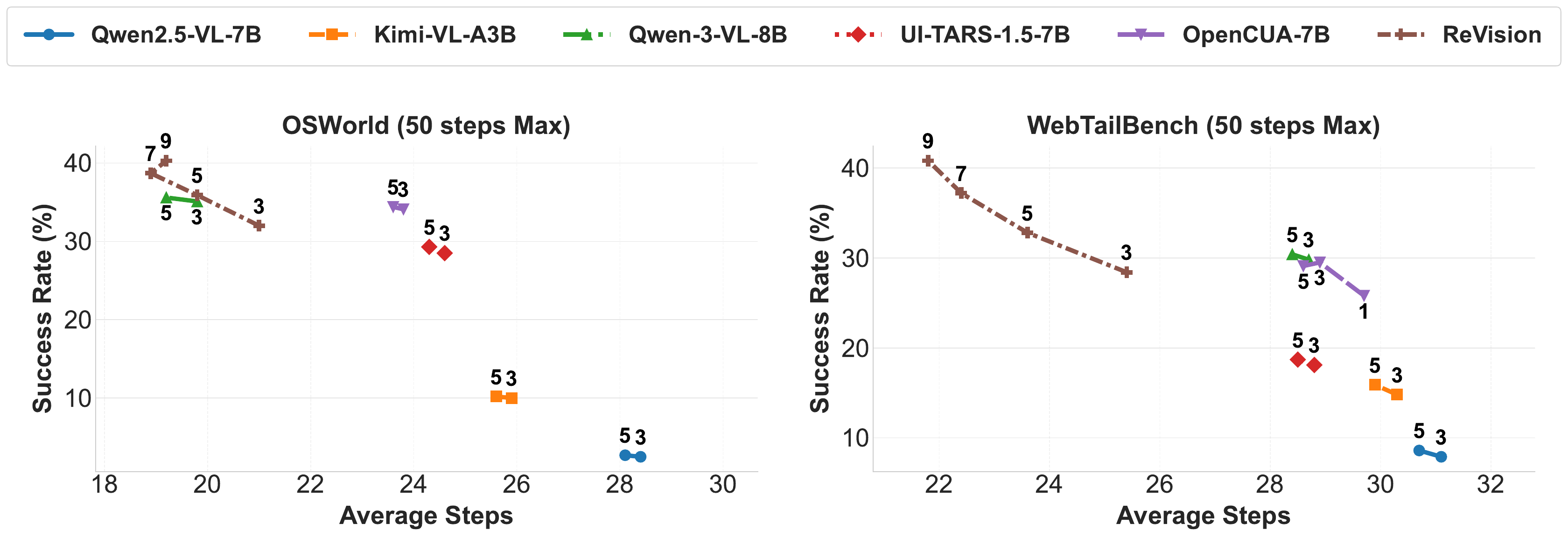}
    \caption{
    Success rate versus average trajectory length (number of steps) for WebTailBench and OSWorld at 50 steps. Detailed numerical results are provided in Tables~\ref{tab:agentnetbench_main} and \ref{tab:webtailbench} in Appendix~\ref{app:main_results_tables}.
    }
    \label{fig:more_steps_vs_sr}
\end{figure*}

Figures~\ref{fig:more_token_vs_sr} and~\ref{fig:more_steps_vs_sr} provide additional views of the efficiency-performance trade-offs across benchmarks and step budgets. Consistent with the main results, \textbf{ReVision} achieves a more favorable trade-off, reaching higher success rates with substantially fewer tokens per step.

\section{Region-Level Grouping and Learned Filtering}
\label{app:omniparser_ablation}

\paragraph{Learned filtering vs. cosine similarity.}
We first analyze the impact of replacing cosine similarity with a lightweight classifier for redundancy detection. Across both Qwen and DINOv2 embeddings, classifier-based filtering consistently improves performance (e.g., Qwen: 72.3 $\rightarrow$ 72.9 SR; DINOv2: 71.7 $\rightarrow$ 72.1), while slightly reducing token usage. This improvement stems from the classifier’s ability to learn an adaptive decision boundary, in contrast to cosine similarity which relies on a fixed threshold (set to 0.95 in our experiments). These results suggest that even simple learned filtering can better capture context-dependent redundancy.

\paragraph{Effect of region-level grouping (OmniParser).}
We next evaluate the effect of incorporating region-level structure using OmniParser. By grouping semantically coherent UI elements, OmniParser enables more structured redundancy detection, leading to improved performance and stronger token reduction (\textbf{74.6 SR} on AgentNetBench and \textbf{35.2 SR@100} on OSWorld, with lower visual token ratios). However, this improvement comes at a significant computational cost. OmniParser introduces substantial latency (\textbf{$\sim$558--572 ms}), which is an order of magnitude higher than embedding-based approaches. In contrast, ReVision (RTS) avoids region parsing at inference time and maintains low latency (\textbf{$\sim$22--23 ms}) while achieving competitive performance (\textbf{73.8 SR}).  These results highlight a key trade-off: while structured region-level grouping can improve redundancy detection, it incurs high overhead that may limit practical deployment. Our approach instead leverages such structure during training, enabling efficient inference without sacrificing performance.

\begin{table*}[t]
    \centering
    \renewcommand{\arraystretch}{1.2}
    \resizebox{\textwidth}{!}{%
    \begin{tabular}{l|cccc|cccc}
        \hline
        \multirow{2}{*}{\textbf{Strategy}} 
        & \multicolumn{4}{c|}{\textbf{AgentNetBench}} 
        & \multicolumn{4}{c}{\textbf{OSWorld}} \\ \cline{2-9} 
        & \textbf{SR} & \textbf{Avg Tok/Step} & \textbf{Vis Tok\%} & \textbf{Latency (ms)} 
        & \textbf{SR@100} & \textbf{Avg Tok/Step} & \textbf{Vis Tok\%} & \textbf{Latency (ms)} \\ 
        \hline

        \rowcolor{gray!20}
        No Drop 
        & 72.5 & 15076 & 92.9 & 0 
        & 32.3 & 15071 & 92.9 & 0 \\ 
        \hline

        Qwen2.5-VL-7B (CosSim) 
        & 72.3 & 9424 & 85.6 & 6 
        & 32.1 & 7624 & 84.1 & 8 \\

        Qwen2.5-VL-7B (Classifier) 
        & 72.9 & 9301 & 85.2 & 10 
        & 32.6 & 7480 & 83.8 & 13 \\

        DINOv2-base (CosSim) 
        & 71.7 & 9682 & 86.2 & 26 
        & 31.4 & 7915 & 84.9 & 31 \\

        DINOv2-base (Classifier) 
        & 72.1 & 9520 & 85.9 & 29 
        & 31.9 & 7768 & 84.5 & 36 \\ 
        \hline

        \rowcolor{cyan!4}
        RTS + OmniParser 
        & 74.6 & 8420 & 79.8 & 558 
        & 35.2 & 6485 & 78.9 & 572 \\

        \rowcolor{cyan!12}
        \textbf{RTS (Ours)} 
        & 73.8 & 8975 & 83.4 & 23 
        & 34.0 & 6963 & 82.9 & 22 \\ 

        \hline
    \end{tabular}
    }
    \caption{\textbf{Ablation on learned filtering and region-level grouping.} We compare cosine similarity and classifier-based filtering across embedding backbones, as well as region-based grouping with OmniParser. Classifier-based methods provide consistent improvements over cosine similarity by learning adaptive thresholds. Incorporating OmniParser further improves performance and reduces visual redundancy, but introduces substantial latency overhead. ReVision (RTS) achieves a strong balance, maintaining competitive performance with significantly lower inference cost.}
    \label{tab:token_selection_appendix}
\end{table*}
\section{Training and Implementation Details}
\label{app:training_details}

\paragraph{Training setup.}
We follow the OpenCUA framework~\citep{wang2025opencuaopenfoundationscomputeruse} and use the same training configuration for fair comparison. All models are trained on a cluster of $8\times$ NVIDIA H200 GPUs. We use standard autoregressive next-token prediction, where the loss is applied only to text tokens (reasoning and actions), while visual tokens are used as conditioning input. For each history window size $k$, we train a separate model using trajectory segments with up to $k$ images, ensuring that the training distribution matches the inference setting.

\paragraph{Data construction.}
Training samples are constructed from AgentNet trajectories by applying a sliding window over interaction sequences. At each step $t$, we form a sample consisting of the most recent $k$ images $\mathcal{I}_{t-k+1:t}$ together with all previous reasoning and actions. This converts trajectories into step-level supervision and increases the number of training samples.

\paragraph{Token filtering.}
During training, we apply the same ReVision token filtering pipeline used at inference time (Algorithm~\ref{alg:revision_inference}). For each pair of consecutive images, RTS compares corresponding patches and produces a binary mask to remove redundant tokens. The first image in each window is kept intact, while subsequent images retain only non-redundant patches. This ensures that the model learns to operate under partially observed visual inputs.

\paragraph{Decoding and evaluation.}
During inference, we use a fixed decoding temperature of $T{=}0.0$ to ensure deterministic behavior and isolate the effect of token filtering. All reported results are averaged over three runs. For WebTailBench, we adopt an LLM-as-a-judge protocol using \textsc{GPT-4o} to evaluate step-level correctness and compute final success rates.

\paragraph{Metrics.}
We report success rate (SR) as the primary metric across all benchmarks. For OSWorld and WebTailBench, we additionally report results under different step budgets to capture long-horizon performance. To better understand efficiency, we also analyze the relationship between success rate, token usage, and the average number of interaction steps required to complete tasks.
\section{Case Study}
\label{app:case_study}

To provide a concrete illustration of how \textbf{ReVision} improves efficiency during sequential decision-making, we present a step-by-step case study on OSWorld in Table~\ref{tab:case_study}. The task requires removing tracking data from Amazon by navigating browser settings and disabling privacy-related options. As shown in the table, consecutive screenshots contain substantial visual overlap, particularly in static UI regions such as navigation bars, menus, and background areas. Without token filtering, all patches are retained across steps, leading to significant redundancy in the visual context.

ReVision addresses this by selectively removing visually redundant patches between consecutive frames while preserving regions that are relevant to the current action. This results in a progressively more compact representation, as seen in the “After Token Removal” column, where large portions of unchanged interface elements are omitted. Importantly, despite this reduction, the agent continues to produce correct actions and coherent reasoning at each step. For example, the model successfully identifies and navigates to browser settings, selects the appropriate privacy options, and proceeds to disable ad personalization, all while operating on a reduced visual context.

This example highlights two key advantages of ReVision: (i) it effectively eliminates temporal redundancy without disrupting spatial alignment or task-relevant information, and (ii) it enables the model to rely on temporally distributed evidence, where previously observed content can be implicitly recalled rather than repeatedly re-encoded. Overall, the case study demonstrates that substantial token savings can be achieved in realistic multi-step interactions without sacrificing decision quality.

\clearpage

\clearpage
\onecolumn

\begin{longtable}{>{\centering\arraybackslash}m{\textwidth}}
\caption{Case study of ReVision on OSWorld.}
\label{tab:case_study} \\

\includegraphics[width=0.85\textwidth,page=1]{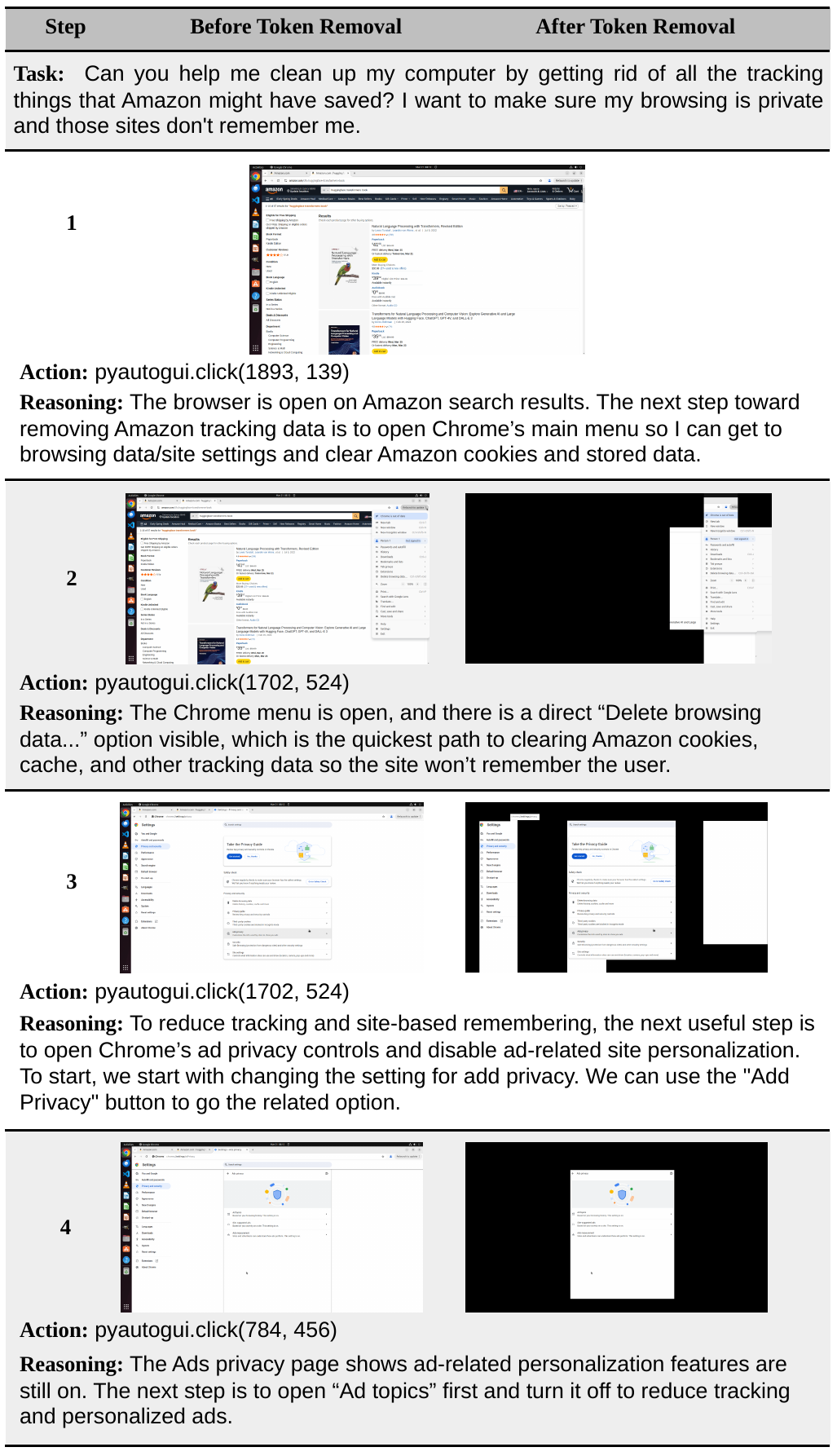} \\[1em]
\includegraphics[width=0.85\textwidth,page=1]{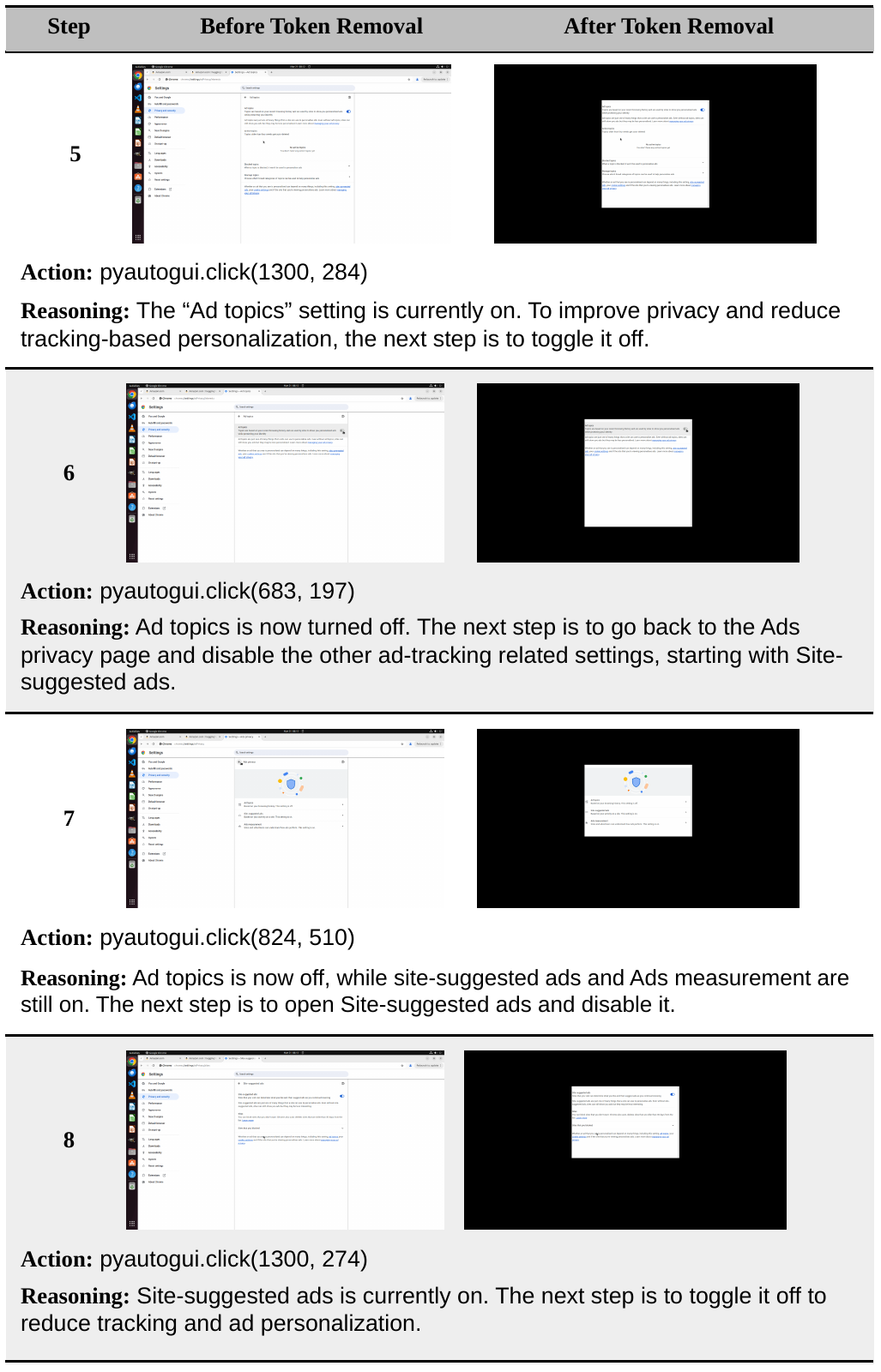} \\[1em]
\includegraphics[width=0.85\textwidth,page=1]{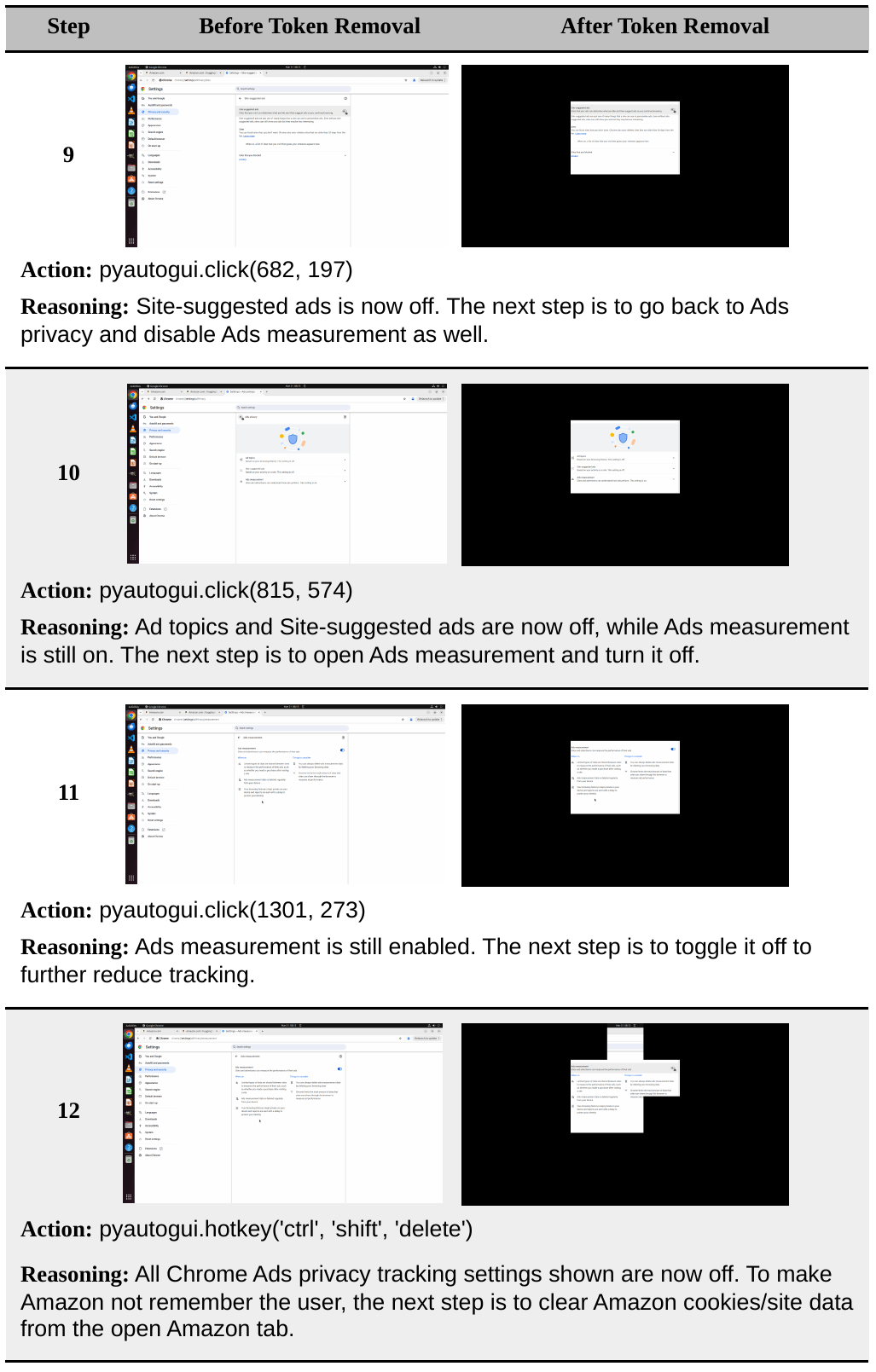} \\[1em]
\includegraphics[width=0.85\textwidth,page=1]{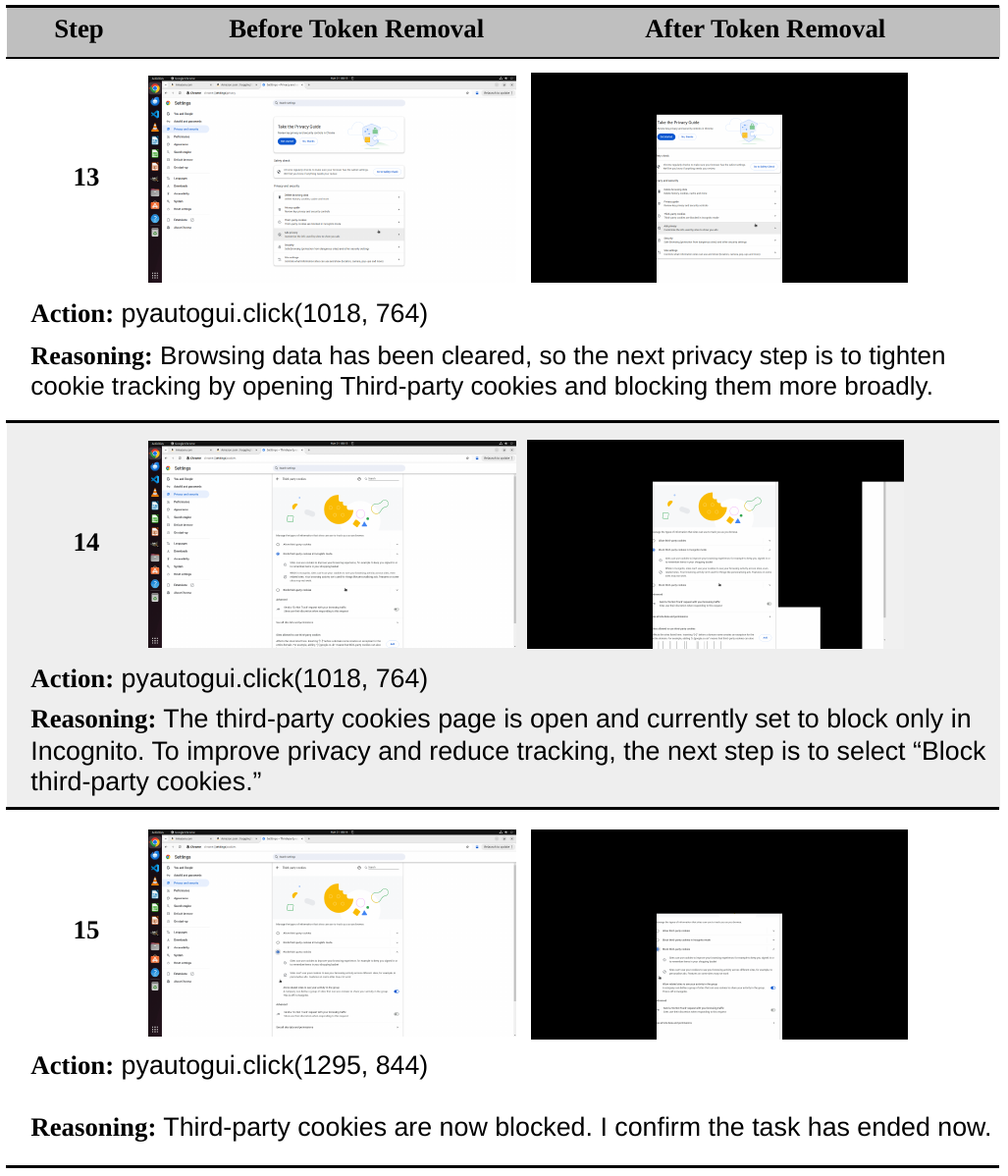} \\[1em]

\end{longtable}

\twocolumn

\end{document}